\documentclass[letterpaper]{article}
\usepackage{uai2019}
\usepackage[margin=1in]{geometry}

\usepackage{times}

\usepackage[english]{babel}
\usepackage{hyperref}

\usepackage{graphicx}

\usepackage[T1]{fontenc}
\usepackage[latin9]{inputenc}
\usepackage{amssymb}
\usepackage{enumerate}

\usepackage{amsthm}
\usepackage{kky}

\newcommand\independent{\protect\mathpalette{\protect\independenT}{\perp}}
\def\independenT#1#2{\mathrel{\rlap{$#1#2$}\mkern2mu{#1#2}}}

\newcommand*{\affaddr}[1]{#1} \newcommand*{\affmark}[1][*]{\textsuperscript{#1}}

\theoremstyle{definition}

\makeatletter
\newcommand{\printfnsymbol}[1]{  \textsuperscript{\@fnsymbol{#1}}}
\makeatother

\title{The Incomplete Rosetta Stone Problem:\\ Identifiability Results for Multi-View Nonlinear ICA}

\author{Luigi Gresele${^{*1,2}}$, Paul K.~Rubenstein${^{*1,3}}$, Arash Mehrjou${^{1,4}}$, Francesco Locatello${^{1,5}}$ and Bernhard Sch\"olkopf${^{1}}$\\
\affaddr{\affmark[1]Empirical Inference, Max Planck Institute for Intelligent Systems, T\"ubingen, Germany.}\\
\affaddr{\affmark[2]Max Planck Institute for Biological Cybernetics, T\"ubingen, Germany.}\\
\affaddr{\affmark[3]Machine Learning Group, University of  Cambridge,  United  Kingdom.}\\
\affaddr{\affmark[4]Max Planck ETH Center for Learning Systems, Z{\"u}rich, Switzerland.}\\
\affaddr{\affmark[5]BMI, Dept. for Computer Science, ETH Z{\"u}rich, Switzerland.}\\
\\\\}

\begin{document}

\maketitle
\begin{NoHyper}
\let\thefootnote\relax\footnote{
$^*$Equal contribution.
}
\end{NoHyper}

\begin{abstract}
We consider the problem of recovering a common latent source with independent components from multiple views.
This applies to settings in which a variable is measured with multiple experimental modalities, and where the goal is to synthesize the disparate measurements into a single unified representation.
We consider the case that the observed views are a nonlinear mixing of component-wise corruptions of the sources.
When the views are considered separately, this reduces to nonlinear Independent Component Analysis (ICA) for which it is provably impossible to undo the mixing.
We present novel identifiability proofs that this is possible when the multiple views are considered jointly,
showing that the mixing can theoretically be undone using function approximators such as deep neural networks.
In contrast to known identifiability results for nonlinear ICA, we prove that independent latent sources with arbitrary mixing can be recovered as long as multiple, sufficiently different noisy views are available.

\end{abstract}

\section{INTRODUCTION}

We consider the setting described by the following generative model    \begin{align}
        \bm{x}_1 &= \bm{f}_1(\bm{s}) \label{eq:nonlinear-ica-1}\\
        \bm{x}_2 &= \bm{f}_2(\bm{s}) \label{eq:nonlinear-ica-2}\\
        p(\bm{s}) &= \prod_{i} p_i(s_i) \label{eq:firstind}\,,
    \end{align}
where $\bm{x}_1, \bm{x}_2, \bm{s} \in \mathbb{R}^D$ and $\bm{f}_1, \bm{f}_2$ are arbitrary smooth and invertible transformations of the latent variable $\bm{s} = (s_1, \ldots, s_D)$ with mutually independent components.
The goal is to recover $\bm{s}$, undoing the mixing induced by the $\bm{f}_i$, in the case where only observations of $\bm{x}_1$ and $\bm{x}_2$ are available.

The two decoupled problems defined by considering pairs of Equations \ref{eq:nonlinear-ica-1}, \ref{eq:firstind} and \ref{eq:nonlinear-ica-2}, \ref{eq:firstind} separately are instances of Independent Component Analysis (ICA). This unsupervised learning method aims at providing a principled approach to disentanglement of independent latent components, blind source separation, and feature extraction~\cite{hyvarinen2000independent}. Its applications are ubiquitous, including neuroimaging~\cite{mckeown1998independent}, signal processing~\cite{sawada2003direction}, text mining~\cite{honkela2010wordica}, astronomy~\cite{nuzillard2000blind} and financial time series analysis~\cite{oja2000independent}.
An ICA problem is identifiable when it is provably possible to simultaneously undo the mixing and recover the sources $\bm{s}$
up to tolerable ambiguities.
Proofs of identifiability are crucial for the characterization of reliable ICA methods; in absence of these, we cannot be confident that a method successfully retrieves the true sources, even within controlled settings.

The case in which $\bm{f}_i$ is a linear function, called linear ICA, has been shown to be identifiable if at most one of the latent components is Gaussian~\cite{darmois1953analyse, skitovich1954linear, comon1994independent}.
This triggered the development of algorithms and encouraged their application. In contrast, the nonlinear ICA problem was shown to be provably unidentifiable without further assumptions on the data generating process~\cite{hyvarinen1999nonlinear}.
Much research in this field has thus attempted to characterize the assumptions under which identifiability holds.
Such assumptions may be grouped into two main categories: (i) those regarding properties of the sources (e.g. non-stationarity or time correlation in time series settings~\cite{cardoso2001three, singer2008non}); and (ii) those restricting the functional form of the mixing functions (e.g., post-nonlinear mixing~\cite{taleb1999source}).

A recent breakthrough was to leverage a technique known as contrastive learning, a method recasting the problem of unsupervised learning as a supervised one~\cite{gutmann2010noise,hyvarinen2016unsupervised, pmlr-v54-hyvarinen17a, hyvarinen19a}. This is a powerful proof technique, which additionally provides algorithms which can be practically implemented using modern deep learning frameworks.
The setup in~\cite{hyvarinen2016unsupervised, pmlr-v54-hyvarinen17a, hyvarinen19a} makes strong assumptions on the data generating mechanism, but allows for arbitrary nonlinear mixing of the sources. However, the
unconditional
independence assumption of the sources (Equation \ref{eq:firstind}) is replaced by a {\em conditional} independence statement, and requires observations of the additional variable conditioned on.

In this paper, we employ contrastive learning to address the setting specified by Equations \ref{eq:nonlinear-ica-1}--\ref{eq:firstind}, where in contrast to~\cite{hyvarinen19a}, no observations of parent variables of the sources are available. This corresponds to cases in which multiple recordings of the same process, acquired with different instruments and possibly different modalities, are available, and the goal is to find an unambiguous representation of the latent state common to all. Multiview settings of this sort are common in large biomedical and neuroimaging datasets~\cite{allen2012uk, miller2016multimodal, van2013wu, shafto2014cambridge}, motivating the need for reliable statistical tools enabling simultaneous handling of multiple sets of variables.

As a metaphor for such a setting, consider the story of the Rosetta Stone, a stele discovered during Napoleon's campaign in Egypt in 1799, inscribed with three versions of a decree issued at Memphis in 196 BC.
The realization that the stone reported the same text translated into three different languages led the French philologist Champollion to succeed in translating two unknown languages (Ancient Egyptian, in hieroglyphic script and Demotic script) by exploiting a known one (Ancient Greek).
Rather, we consider the radically unsupervised task in which, given a Rosetta Stone with only two texts, both in unknown languages, we want to learn an unambiguous common representation for both of them.

The main contribution of this paper is to show that jointly addressing multiple demixing problems allows for identifiability with assumptions which do not directly refer to the sources, nor to restriction of the class of mixing functions, but rather to the conditional probability distribution of one observation given the other.
This provides identifiability results in a novel setting, with assumptions entailing a different interpretation - namely, that the views have to be sufficiently diverse.

The remainder of this paper is organized as follows.
In Section \ref{sec:nlica} we provide background information about the technique of contrastive learning for ICA and briefly review recent work that employs it.
In Section~\ref{sec:nlica-with-multiple-views} we present our main results, providing identifiability for different multi-view settings.
In Section~\ref{sec:related-work} we discuss other relevant works in the literature.
Finally, we summarize and discuss our results in Section~\ref{sec:on_suffistv}.

\section{NONLINEAR ICA WITH CONTRASTIVE LEARNING}
\label{sec:nlica}

Consider the nonlinear ICA setting, where observations of a variable $\bm{x} = \bm{f}(\bm{s})$ are available, where $\bm{f}$ is an arbitrary nonlinear invertible mixing. The proof of non-identifiability for the general case with unconditionally independent sources was an important negative result~\cite{hyvarinen1999nonlinear}. We review it briefly in Appendix \ref{sec:on_unident}.

A proposed modification of this setting~\cite{hyvarinen19a} involves an auxiliary observed variable $\bm{u}$ and a change in the independence properties. If the \textit{unconditional independence} is substituted with a \textit{conditional independence} given the auxiliary variable $\bm{u}$, i.e.
\begin{equation}
    \log p( \bm{s} | \bm{u}) = \sum_i q_i(s_i, \bm{u})\,,\label{eq:secondind}
\end{equation}
for some functions $q_i$, the model becomes identifiable. The conditional independence statement in Equation \ref{eq:secondind} can be interpreted as positing that $\bm{u}$ is a parent of the sources $\bm{s}$. A further assumption on the effect of variations in $\bm{u}$ on $\bm{x}$, called \textit{variability} in the paper, is required. Intuitively, it demands that $\bm{u}$ has a sufficiently diverse influence on $\bm{x}$.

In the setting described above, a constructive proof of identifiability is attained by exploiting contrastive learning~\cite{gutmann2010noise}.\\\label{sec:contr_learn}
This technique transforms a density ratio estimation problem into one of supervised function approximation. This idea has a long history~\cite{friedman2001elements}, and has attracted attention in machine learning in recent years~\cite{goodfellow2014generative, gutmann2010noise}. We recapitulate the method in Appendix \ref{sec:converged}.

In the setting of nonlinear ICA with auxiliary variables, contrastive learning can be exploited by training a classifier to distinguish between a tuple sampled from the joint distribution, which we denote as $(\bm{x}, \bm{u})$, and one where $\bm{u}^*$ is a sample generated from the marginal $p(\bm{u})$ independently of $\bm{x}$, $(\bm{x}, \bm{u}^*)$.
Intuitively, tuples drawn from the former distribution correspond to the same sources $\bm{s}$, and thus share information, while tuples from the latter correspond to different sources and thus do not share information.
Since the marginals of both distributions are equal, the classifier must learn to distinguish between them based on the common information shared by $\bm{x}$ and $\bm{u}$; that is, ultimately, $\bm{s}$.

With this method, the reconstruction of $\bm{s}$ is only possible up to an invertible scalar ``gauge'' transformation. This is due to a fundamental ambiguity in the setup of nonlinear ICA and does not represent a limitation of their results; it can therefore be considered a trivial one. We further comment on this in Appendix \ref{sec:gauge}.

\section{NONLINEAR ICA WITH MULTIPLE VIEWS}
\label{sec:nlica-with-multiple-views}

We described how naively splitting Equations \ref{eq:nonlinear-ica-1}, \ref{eq:nonlinear-ica-2} and \ref{eq:firstind} into two separate nonlinear ICA problems renders both problems non-identifiable, unless strong assumptions are made on the $\bm{f}_i$ or the distribution of $\bm{s}$.

In the Rosetta stone story, awareness that different texts reported on the stele were linked by a common topic helped solving the translation problem; similarly, in our setting, matched observations of the two views are linked through the shared latent variable $\bm{s}$. Thus the central question we investigate is whether these assumptions can be relaxed by exploiting the structure of the generative model; that is, whether jointly observing $\bm{x}_1$ and $\bm{x}_2$ provides sufficient constraints to the inverse problem, thus removing the ambiguities present in the vanilla nonlinear ICA setting.
We consider a contrastive learning task in which a classifier is trained to distinguish between pairs $(\bm{x}_1, \bm{x}_2)$ corresponding to the same $\bm{s}$ and $(\bm{x}_1, \bm{x}^*_2)$ corresponding to different realizations of $\bm{s}$.
As discussed in Section \ref{sec:nlica}, the classifier will be forced to employ the information shared by the simultaneous views in order to distinguish the two classes.
As we show, this ultimately results in recovering $\bm{s}$ (up to unavoidable ambiguities).

For technical reasons discussed in Appendix
\ref{sec:converged}, our method requires some stochasticity in the relationship between $\bm{s}$ and at least one of the $\bm{x}_i$.
However this is not a significant constraint in practice; in most real settings observations are corrupted by noise, and a truly deterministic relationship between $\bm{s}$ and the $\bm{x}_i$ would be unrealistic.
We will consider a component-wise independent corruption of our sources, i.e.~$\bm{x}_1 = \bm{f}_1 \circ \bm{g}_1(\bm{s}, \bm{n}_1)$ with $g_{1i}(\bm{s}, \bm{n}_1) = g_{1i}(s_i, n_{1i})$, where the components of $\bm{n}_{1}$ are mutually independent, and similar for $\bm{x}_2$. The noise variables $\bm{n}_1$, $\bm{n}_2$ and the sources $\bm{s}$ are assumed to be mutually independent.
Note that this only puts constraints on the way the signal is corrupted by the noise, namely $\bm{g}$, and not on the mixing $\bm{f}$.
We will refer to such $\bm{g}$ as \emph{component-wise corrupter} throughout, and to its output as \emph{corruption}.
In the the vanilla ICA setting, inverting the mixing and recovering the sources $\bm{s}$ are equivalent; in the setting that we consider, the inversion of the mixing $\bm{f}$ only implies recovering the sources up to the effect of the corrupter $\bm{g}$.

We will consider three instances of the general setting, providing identifiability results for each.
\begin{enumerate}[I.]
    \item First we consider the case that only one of the observations, $\bm{x}_2$, is corrupted with noise. This corresponds, for instance, to a setting in which one accurate measurement device is supplemented with a second noisy device. We show that in this setting it is possible to fully reconstruct $\bm{s}$ using the noiseless variable (Section \ref{sec:onenoisless}).
    \item Next, we consider the case that both variables are corrupted with noise. In this setting, it is possible to recover $\bm{s}$ up to the corruptions.     Furthermore, we show that $\bm{s}$ can be recovered with arbitrary precision in the limit that the corruptions go to zero (Section \ref{sec:constrained}).
    \item Finally, we consider the case of having $N$ simultaneous views of the source $\bm{s}$ rather than just two.
    When considering the limit $N \rightarrow \infty$, we prove sufficient conditions under which it is possible to reconstruct $\bm{s}$ even if each observation is corrupted by noise (Section \ref{sec:multiple}).
\end{enumerate}

To the best of our knowledge, no result of identifiability of latent sources in the case in which only corrupted, mixed versions are observed has been given before.

\subsection{ONE NOISELESS VIEW}
\label{sec:onenoisless}
\begin{figure}[t!]
    \centering
    \includegraphics[scale=0.3]{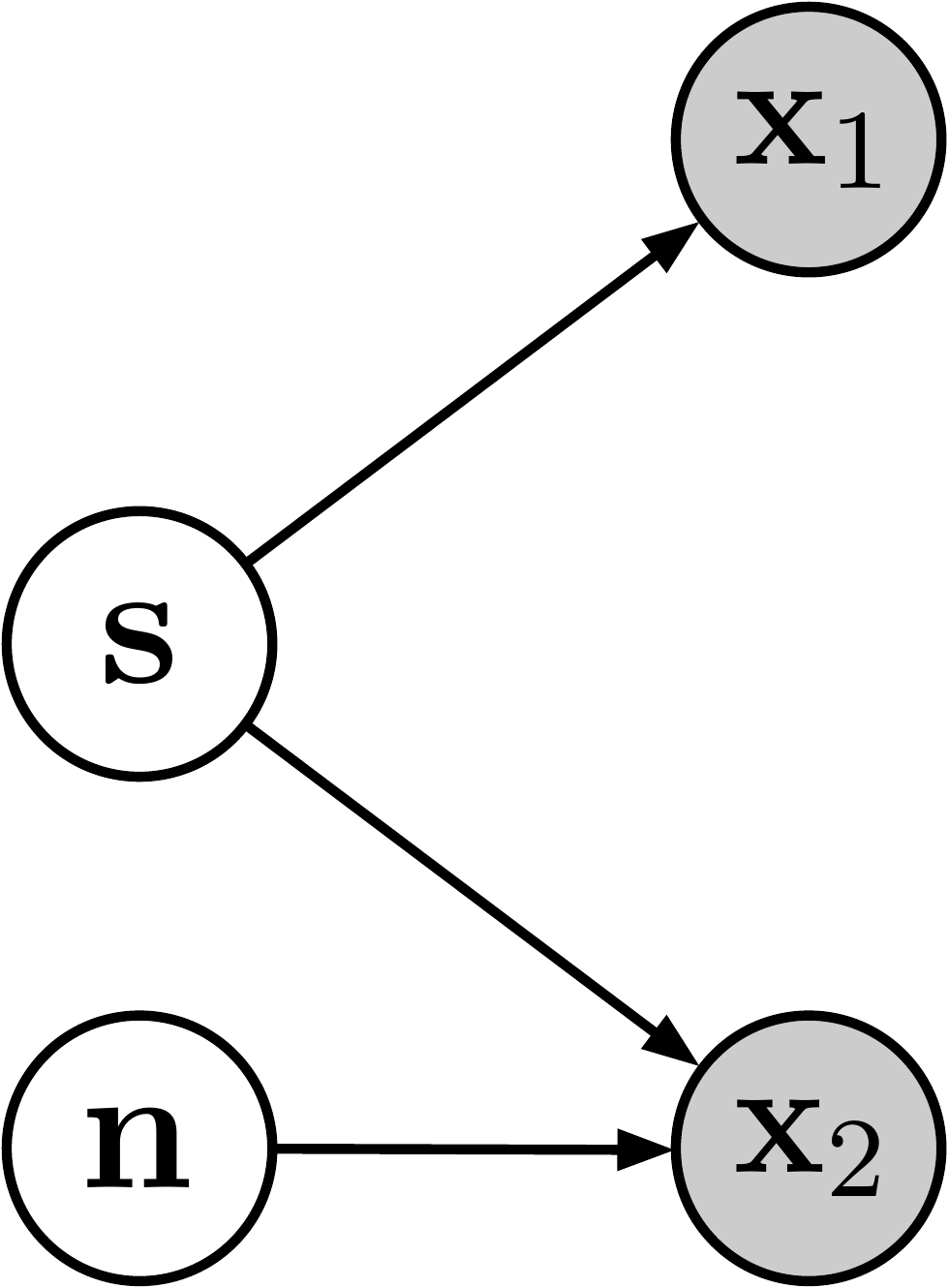}
    \caption{The setting considered in Section \ref{sec:onenoisless}. Two views of the sources are available, one of which, $\bm{x}_1$, is not corrupted by noise. In this and all other figures, each node is a deterministic function of all its parents in the graph.
        }
    \label{fig:generalized_hsr_basic}
\end{figure}

Consider the generative model
\begin{align}
\bm{x}_{1}&=\bm{f}_{1}(\bm{s}) \label{eq:sem2_1}\\
\bm{x}_{2}&=\bm{f}_{2}(\bm{g}(\bm{s}, \bm{n})) \label{eq:sem2_2} \\
p(\bm{s}) &= \prod_{i} p_i(s_i) \nonumber \\
p(\bm{n}) &= \prod_{i} p_i(n_i) \label{eq:indep}
\end{align}
where $\bm{f}_1$ and $\bm{f}_2$ are invertible, $\bm{g}$ is a component-wise corrupter, $\bm{n} \independent \bm{s}$ and $\bm{x}_1$ and $\bm{x}_2$ are observed.
This is represented in Figure \ref{fig:generalized_hsr_basic}.

Subject to some assumptions, it is possible to recover $\bm{s}$ up to the
component-wise invertible ambiguity.

\begin{theorem}
\label{thm:noiseless1}
The difference of the log joint probability and log product of marginals of the observed variables in the generative model specified by Equations \ref{eq:sem2_1}-\ref{eq:indep} admits the following factorisation:
\begin{align}
&\log p(\bm{x}_1, \bm{x}_2) - \log p(\bm{x}_1) p(\bm{x}_2) \nonumber \\
&= \log p(\bm{x}_2 | \bm{x}_1) - \log p(\bm{x}_2) \nonumber\\
&= \left(\sum_i \alpha_i(s_{i}, g_i(s_i, n_i)) + \log \det J \right) \nonumber\\
&\qquad - \left( \sum_i \delta_i(g_i(s_i, n_i)) + \log \det J\right) \nonumber\\
&= \sum_i \alpha_i(s_{i}, g_i(s_i, n_i)) - \sum_i \delta_i(g_i(s_i, n_i))\label{eq:logdens_noiesless_1} \,
\end{align}
where $s_i=f^{-1}_{1i}(\bm{x}_1)$, $g_i=f^{-1}_{2i}(\bm{x}_2)$,
and $J$ is the Jacobian of the transformation $f^{-1}_2$ (note that the introduced Jacobians cancel).
Suppose that
\begin{enumerate}
    \item $\bm{\alpha}$ satisfies the \emph{Sufficiently Distinct Views} assumption (see after this theorem).
    \item We train a classifier to discriminate between
\begin{align*}
(\bm{x}_{1},\bm{x}_{2}) \text{ vs. } (\bm{x}_{1},\bm{x}_{2}^{*})\,,
\end{align*}
where $(\bm{x}_{1},\bm{x}_{2})$ correspond to the same realization of $\bm{s}$ and $(\bm{x}_{1},\bm{x}_{2}^{*})$ correspond to different realizations of $\bm{s}$.
\item The classifier is constrained to use a regression function of the form
\begin{equation*}
r(\bm{x}_{1},\bm{x}_{2})=\sum_{i}\psi_{i}(h_{i}(\bm{x}_{1}),\bm{x}_{2})
\end{equation*}
where $\bm{h} =(h_{1}, \ldots, h_{n})$  are invertible, smooth and have smooth inverse.
\end{enumerate}

Then, in the limit of infinite data and with universal approximation capacity, $\bm{h}$ inverts $\bm{f}_1$ in the sense that the $h_{i}(\bm{x}_1)$ recover the independent  components of $\bm{s}$ up to component-wise invertible transformations.
\end{theorem}
The proof can be found in Appendix \ref{appendix:proof-thm1}.
The assumption of invertibility for $\bm{h}$ could be satisfied by, e.g., the use of normalizing flows~\cite{rezende2015variational, chen2018neural} or deep invertible networks~\cite{jacobsen_hal-01712808}.

We remark that at several points in this paper we consider the difference between two log-probabilities.
In all of these cases, the Jacobians introduced by a change of variables cancel out as in Equation \ref{eq:logdens_noiesless_1}.
For brevity we omit explanation of this fact in the rest of the results.

The \emph{Sufficiently Distinct Views (SDV)} assumption specifies in a technical sense that the two views available are sufficiently different from one another,
resulting in more information being available in totality than from each view individually.
In the context of Theorem~\ref{thm:noiseless1}, it is an assumption about the log-probability of the \emph{corruption} conditioned on the source.
Informally, it demands that the probability distribution of the corruption should vary significantly as a result of conditioning on different values of the source.

\begin{definition}[Sufficiently Distinct Views]\label{suff_dist_assumption}
Let $\alpha_i(y_i, t_i)$, $i=1,\ldots, N$ be functions of two arguments.
Denote by $\bm\alpha$ the vector of functions and define
\begin{align}
\alpha'_{i}(y_i, t_i)&= \partial \alpha_{i}(y_i, t_i)/\partial t, \label{eq:convention1}\\
\alpha''_{i}(y_i, t_i)&=\partial^2 \alpha_{i}(y_i, t_i)/\partial t^2\, \label{eq:convention2}\\
\bm{w}_{\bm\alpha}(\bm{y}, \bm{t}) &= (\alpha''_{1}, \ldots, \alpha''_{D}, \alpha'_{1}, \ldots,\alpha'_{D}).
\end{align}
We say that $\bm{\alpha}$ satisfies the assumption of \emph{Sufficiently Distinct Views (SDV)} if for any value of $\bm{y}$, there exist $2D$ distinct values $\bm{t}_j$, $j=1, \ldots, 2D$ such that the vectors $\bm{w}(\bm{y},\bm{t}_j)$ are linearly independent.
    \\    \end{definition}
This is closely related to the Assumption of Variability in~\cite{hyvarinen19a}.
We provide simple cases of conditional log-probability density functions satisfying and violating the SDV assumption in Appendix \ref{appendix:sdv}.

Theorem \ref{thm:noiseless1} shows that by jointly considering the two views, it is possible to recover $\bm{s}$, in contrast to the single-view setting.
This result can be extended to learn the inverse of $\bm{f}_2$ up to component-wise invertible functions.
\begin{corollary}
\label{crl:noiseless1}
Consider the setting of Theorem \ref{thm:noiseless1}, and the alternative factorization of the log joint probability given by
\begin{align}
&\log p(\bm{x}_1, \bm{x}_2) - \log p(\bm{x}_1) p(\bm{x}_2) \nonumber \\
&= \log p(\bm{x}_1 | \bm{x}_2) - \log p(\bm{x}_1)\nonumber \\
&= \sum_i \gamma_i(s_{i}, g_i(s_i, n_i)) - \sum_i \beta_i(s_i)) \label{eq:logdens_noiesless_2}\,.
\end{align}
Suppose that $\bm{\gamma}$ satisfies the SDV assumption.
Replacing the regression function with
\begin{equation*}
r(\bm{x}_{1},\bm{x}_{2})=\sum_{i}\psi_{i}(\bm{x}_{1}, h_{i}(\bm{x}_{2}))
\end{equation*}
results in $\bm{h}$ inverting $\bm{f}_2$ in the sense that the $h_{i}(\bm{x}_2)$ recover the independent components of $\bm{g}(\bm{s}, \bm{n})$ up  to component-wise invertible transformations.
\end{corollary}The proof can be found in Appendix \ref{appendix:proof-cor2}.
These two results together mean that it is possible to learn inverses $\bm{h}_1$ and $\bm{h}_2$ of $\bm{f}_1$ and $\bm{f}_2$, and therefore to recover $\bm{s}$ and $\bm{g}(\bm{s}, \bm{n})$, up to component-wise intertible functions.
Note, however, that doing so requires running two separate algorithms.
Furthermore, there is no guarantee that the learned inverses $\bm{h}_1$ and $\bm{h}_2$ are `aligned' in the sense that for each $i$ the components $\bm{h}_{1i}(\bm{x}_1)$ and $\bm{h}_{2i}(\bm{x}_2)$ correspond to the same components of $\bm{s}$.

This problem of misalignment can be resolved by changing the form of the regression function.

\begin{theorem}\label{thm:demixing}
Consider the settings of Theorem \ref{thm:noiseless1} and Corollary \ref{crl:noiseless1}.
Suppose that both $\bm{\alpha}$ and $\bm{\gamma}$ satisfy the SDV assumption.
Replacing the regression function with
\begin{equation}\label{eqn:double-regression-fn}
r(\bm{x}_{1},\bm{x}_{2})=\sum_{i}\psi_{i}(h_{1,i}(\bm{x}_{1}),h_{2,i}(\bm{x}_{2}))
\end{equation}
results in $\bm{h}_1$, $\bm{h}_2$ inverting $\bm{f}_1$, $\bm{f}_2$ in the sense that the $h_{1,i}(\bm{x}_1)$ and $h_{2,i}(\bm{x}_2)$ recover the independent components of $\bm{s}$ and $\bm{g}(\bm{s}, \bm{n})$ up to two different component-wise invertible transformations. Furthermore, the two representations are aligned, i.e. for $i\not=j$,
\begin{equation*}
    h_{1,i}(\mathbf{x}_{1})\independent h_{2,j}(\mathbf{x}_{2})
\end{equation*}
\end{theorem}
The proof can be found in Appendix \ref{appendix:thm1}.

Note that Theorem \ref{thm:demixing} is \emph{not} a generalisation of Theorem \ref{thm:noiseless1} or Corollary \ref{crl:noiseless1}, since it makes stricter assumptions by imposing the SDV assumption on both $\bm{\alpha}$ and $\bm{\gamma}$.
In contrast, Theorem \ref{thm:noiseless1} and Corollary \ref{crl:noiseless1} require that only one is valid for each.

For cases in which finding aligned representations for $\bm{s}$ and $\bm{g}(\bm{s}, \bm{n})$ are desired, Theorem \ref{thm:demixing} should be applied.
If the only goal is recovery of $\bm{s}$, the assumptions of Theorem \ref{thm:noiseless1} are simpler to verify.

In practical applications, the multi-view scenario is useful in multimodal datasets where one of the two acquisition modalities has much higher signal to noise ratio than the other one (e.g., in neuroimaging, when simultaneous fMRI and Optical Imaging recordings are compared). In such cases, jointly exploiting the multiple modalities would help to discern a meaningful and identifiable latent representation which could not be attained through analysis of the more reliable modality alone.

\subsubsection{Equivalence with Permutation Contrastive Learning for Time Dependent Sources}
Note that the analysis of Theorem~\ref{thm:noiseless1} covers the case of temporally dependent stationary sources analyzed in~\cite{pmlr-v54-hyvarinen17a}.
Indeed, if it is further assumed that $\bm{s}$ and $\bm{g}(\bm{s}, \bm{n})$ are uniformly dependent~\cite{pmlr-v54-hyvarinen17a}, they can be seen as a pair of subsequent time points of an ergodic stationary stochastic process for which the analysis of Theorem 1 of~\cite{pmlr-v54-hyvarinen17a} would hold. In other words, we can define a stochastic process as $p(\bm{s}_{t+1}| \bm{s}_t) := p(\bm{g}(\bm{s}, \bm{n})| \bm{s})$.
Note that while the two formulations are theoretically equivalent, our view offers a wider applicability as it covers the asynchronous sensing of $\bm{s}$, provided that multiple measurements (i.e. $\bm{x}_1, \bm{x}_2$) are available; additionally, our \textit{Sufficiently Distinct Views} assumption does not necessarily imply uniform dependency. Furthermore, while~\cite{pmlr-v54-hyvarinen17a} considers a generative model of the form $\bm{x}(t) = \bm{f}(\bm{s}(t))$, thus constraining the mixing function to be the same for any two data points $\bm{x}(t_1)$, $\bm{x}(t_2)$, in our setting we consider two different mixing functions, $\bm{f}_1$ and $\bm{f}_2$, for the two different views.
Finally, we study this setting as an intermediate step for the following two sections, in which no deterministic function of the sources is observed, learning to invert any of the $\bm{f}_i$ can only recover $\bm{s}$ up to the corruption operated by $\bm{g}$.

\subsection{TWO NOISY VIEWS}
\label{sec:constrained}

\begin{figure}[t!]
    \centering
    \includegraphics[scale=0.3]{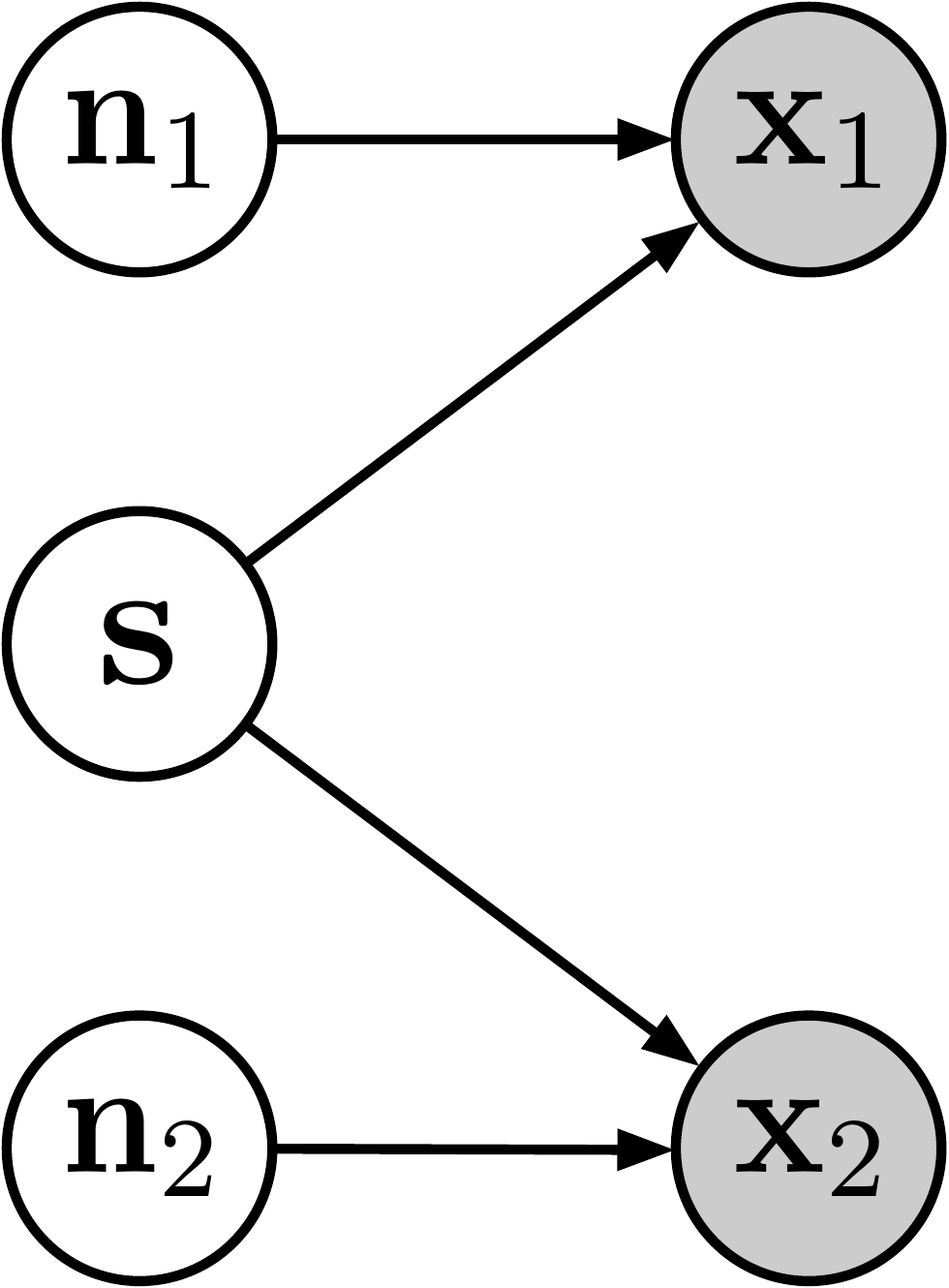}
    \caption{Setting with two views of the sources $\bm{s}$, both corrupted by noise.}
    \label{fig:classic_hsr}
\end{figure}

We next consider the setting in which both variables are corrupted by noise.
Consider the following generative model (represented in Figure \ref{fig:classic_hsr}):
\begin{align*}
\bm{x}_{1}&=\bm{f}_{1}(\bm{g}_{1}(\bm{s},\bm{n}_{1}))  \\
\bm{x}_{2}&=\bm{f}_{2}(\bm{g}_{2}(\bm{s},\bm{n}_{2}))  \,,
\end{align*}
where all variables take value in $\mathbb{R}^D$, and $\bm{f}_{1}$ and $\bm{f}_{2}$ are nonlinear, invertible, deterministic functions,
$\bm{g}_{1}$ and $\bm{g}_{2}$ are component-wise corrupters, and $\bm{s}$ and the $\bm{n}_i$ are independent with independent components.
This class of models generalizes the setting of Section \ref{sec:onenoisless} since by taking $\bm{g}_1(\bm{s}, \bm{n}_1) = \bm{s}$ we reduce to the case of one noiseless observation.

The difference $\log p(\bm{x}_1, \bm{x}_2) - \log p(\bm{x}_1)p(\bm{x}_2)$ admits similar factorizations to those given in Equations \ref{eq:logdens_noiesless_1} and \ref{eq:logdens_noiesless_2}:
\begin{align}
&\log p(\bm{x}_1, \bm{x}_2) - \log p(\bm{x}_1) p(\bm{x}_2) \nonumber\\
&= \log p(\bm{x}_1 | \bm{x}_2) - \log p(\bm{x}_1)\nonumber\\
&= \sum_i \eta_i(g_{1i}(s_i, n_{i1}), g_{2i}(s_i, n_{2i})) - \sum_i \theta_i(g_{1i}(s_i, n_{1i}) \label{eq:noisylogdens_1}\\
&= \log p(\bm{x}_2 | \bm{x}_1) - \log p(\bm{x}_2) \nonumber\\
&= \sum_i \lambda_i(g_{2i}(s_i, n_{2i}), g_{1i}(s_i, n_{1i})) - \sum_i \mu_i(g_{2i}(s_i, n_{2i})) \label{eq:noisylogdens_2}
\end{align}
Since we only have access to corrupted observations, exact recovery of $\bm{s}$ is not possible.
Nonetheless, a generalization of Theorem \ref{thm:demixing} holds showing that the $\bm{f}_i$ can be inverted and $\bm{s}$ recovered up to the corruptions induced by the $\bm{n}_i$ via $\bm{g}_i$.
\begin{theorem}\label{thm:two-noisy-views}
Suppose that $\bm{\eta}$ and $\bm{\lambda}$ satisfy the SDV assumption.
The algorithm described in Theorem \ref{thm:noiseless1} with regression function specified in Equation \ref{eqn:double-regression-fn} results in $\bm{h}_1$ and $\bm{h}_2$ inverting $\bm{f}_1$ and $\bm{f}_2$ in the sense that the $h_{1,i}(\bm{x}_1)$ and $h_{2,i}(\bm{x}_2)$ recover the independent components of $\bm{g}_1(\bm{s}, \bm{n}_1)$ and $\bm{g}_2(\bm{s}, \bm{n}_2)$ up to two different component-wise invertible transformations. Furthermore, the two representations are aligned, i.e. for $i\not=j$,
\begin{equation*}
    h_{1,i}(\mathbf{x}_{1})\independent h_{2,j}(\mathbf{x}_{2})
\end{equation*}
\end{theorem}
The proof can be found in Appendix \ref{appendix:thm1}.

We can thus recover the common source $\bm{s}$ up to the corruptions $\bm{g}_i(\bm{s}, \bm{n}_i)$.
In the limit of the magnitude of one of the noise variables going to zero, the reconstruction of the sources $\sbm$ attained through the corresponding view is exact up to the component-wise invertible functions, as stated in the following corollary.

\begin{corollary}
\label{crl:lownoise}
Let $\bm{n}_1^{(k)} = \frac{1}{k} \cdot  \Tilde{\bm{n}}$ for $k \in \NN$, where $\Tilde{\bm{n}}\in\mathbb{R}^D$ is a fixed random variable, and $\bm{n}_2$ be a random variable that does not depend on $k$.
Let $\bm{h}_1^{(k)}, \bm{h}_2^{(k)}$ be the output of the algorithm specified by Theorem \ref{thm:two-noisy-views} with noise variables $\bm{n}_1^{(k)}$ and $\bm{n}_2$.

Suppose that the corrupters $\bm{g}_i$ satisfy the following two criteria:
\begin{enumerate}[i)]
    \item $\exists \bm{a}  \in \mathbb{R}_{> 0}^D \: $   s.t. $\: \left|\frac{\partial \bm{g}_1(\bm{s},\bm{n})}{\partial \bm{n}} \right|_{\bm{n}=0} \leq \bm{a} \: $ for all $\bm{s}$
    \item $\exists \bm{b}  \in \mathbb{R}_{> 0}^D \: $ s.t. $\: 0<\frac{\partial \bm{g}_1(\bm{s},0)}{\partial \bm{s}} \leq \bm{b}$
\end{enumerate}
Then, denoting by $\bm{E}$ the set of all scalar, invertible functions, we have that
\[
\lim_{k \to \infty} \inf_{\bm{e}\in \bm{E}} \left \|\bm{s} - \bm{e}(\bm{h}_1^{(k)}(\bm{x}_1)) \right \| = 0
\]
\end{corollary}
The proof can be found in Appendix \ref{appendix:thm2}.

Corollary \ref{crl:lownoise} implies that in the limit of small noise, the sources $\bm{s}$ can be recovered exactly.
Condition \textit{i)} upper bounds the influence of $\bm{n}$ on the corruption: we can not hope to retrieve $\bm{s}$ if $\bm{g}(\bm{s}, \bm{n})$ contains too little signal.
Condition \textit{ii)} ensures that the function $\bm{g}$ is invertible with respect to $\bm{s}$ when $\bm{n}$ is equal to zero.
If this were not satisfied, some information about $\bm{s}$ would be washed out by $\bm{g}$ even in absence of noise.
This would make recovery of $\bm{s}$ trivially impossible.

\subsection{MULTIPLE NOISY VIEWS}
\label{sec:multiple}

The results of Section \ref{sec:constrained} state that in the two noisy view setting, $\bm{s}$ can be recovered up to the corruptions.
In the limit that the magnitude of the noises goes to zero, the uncorrupted $\bm{s}$ can be recovered.
The intuition is that the less noise there is, the more information each observation provides about $\bm{s}$.

\begin{figure}[t!]
    \centering
    \includegraphics[scale=0.3]{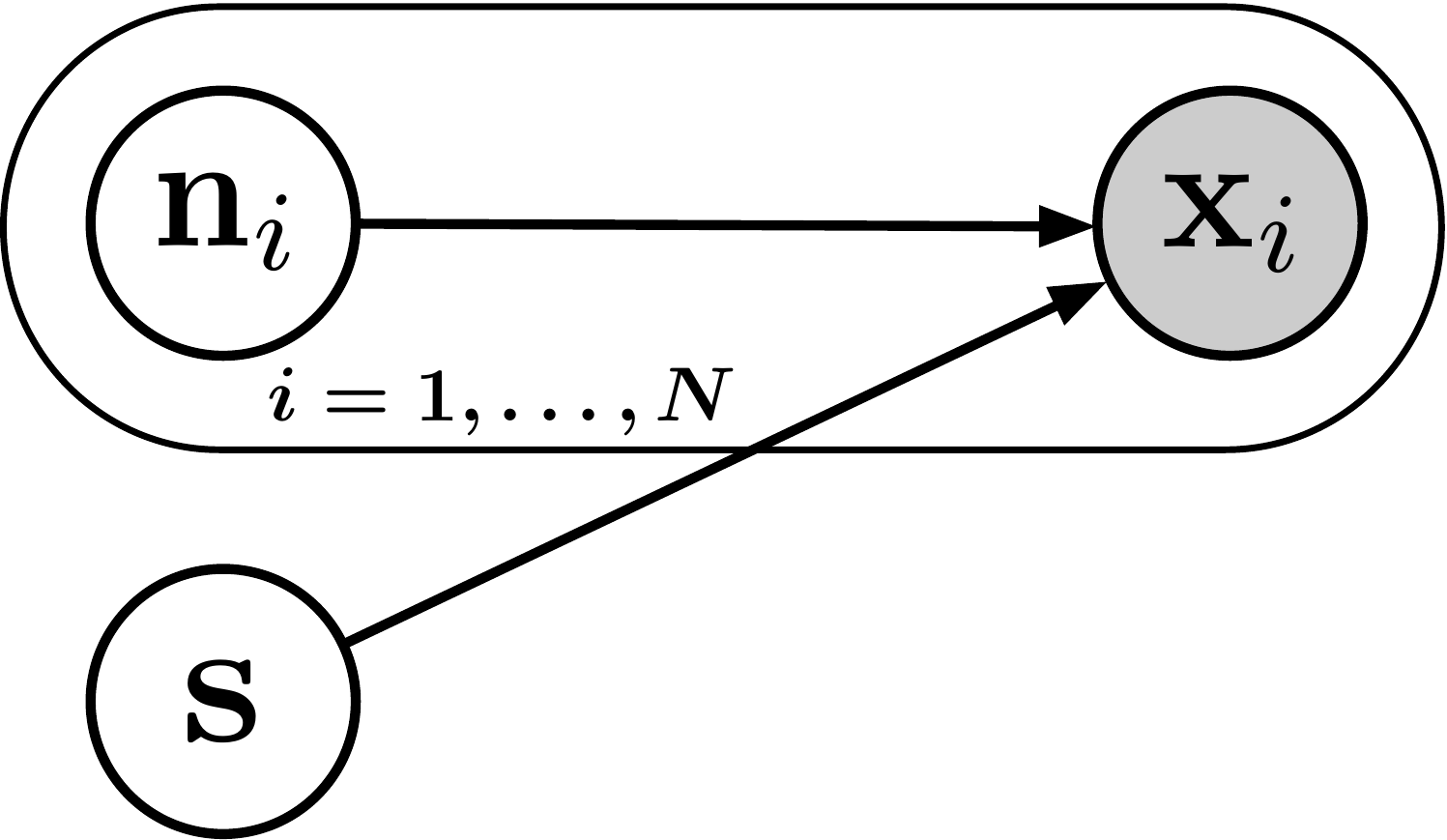}
    \caption{Setting with $N$ corrupted views of the sources.}
    \label{fig:generalized_hsr_many}
\end{figure}

In this section we consider the multi-view setting, where $N$ distinct noisy views of $\bm{s}$ are available,
\begin{equation*}
\bm{x}_{i}=\bm{f}_{i}(\bm{g}_{i}(\bm{s},\bm{n}_{i}))\,\,\,,i=1, \ldots, N\,, \label{eq:multi}\\
\end{equation*}
and the noise variables $\bm{n}_{i}$ are mutually independent, as represented in Figure \ref{fig:generalized_hsr_many}.
Since each view provides additional information about $\bm{s}$, we ask: in the limit as $N \to \infty$, is it possible to reconstruct $\bm{s}$ exactly?

By applying Theorem \ref{thm:two-noisy-views} to the pair $(\bm{x}_1,\bm{x}_i)$ it is possible to recover  $(\bm{g}_1(\bm{s},\bm{n}_1),\bm{g}_i(\bm{s},\bm{n}_i))$ such that the components are aligned, but up to different component-wise invertible functions $\bm{k}_1$ and $\bm{k}_i$.
Running the algorithm on a different pair  $(\bm{x}_1,\bm{x}_{j})$ will result in recovery up to different component-wise invertible functions $\bm{k}'_1$ and $\bm{k}'_j$.

Note that these will \emph{not} necessarily result in  $\bm{k}_i\circ\bm{g}_i(\bm{s},\bm{n}_i)$ and $\bm{k}'_j\circ\bm{g}_j(\bm{s},\bm{n}_j)$ being aligned with each other.
However, the components of $\bm{k}_1\circ\bm{g}_1(\bm{s},\bm{n}_1)$ and $\bm{k}'_1\circ\bm{g}_1(\bm{s},\bm{n}_1)$ are the same, up to permutation and component-wise invertible functions.
This permutation can therefore be undone by performing independence testing between each pair of components.
Components that are `different' will be independent; those that are the same will be deterministically related.
Therefore, they can be used as a reference to permute the components of $\bm{k}'_j$ and make it aligned with $\bm{k}_i$.

The problem is then how to combine the information from each aligned $\bm{k}_i \circ \bm{g}_i(\bm{s},\bm{n}_i)$ to more precisely identify $\bm{s}$.
The fact that the components are recovered up to \emph{different} scalar invertible functions makes combining information from different views non-trivial.

As a first step in this direction, we consider the special case that each $\bm{g}_i$ acts additively and each $\bm{n}_i$ is zero mean and each of $\bm{s}$ and the $\bm{n}_i$ are independent with independent components.
\begin{align}
\left.
   \begin{array}{ll}
&\bm{x}_{i}=\bm{f}_{i}(\bm{s} + \bm{n}_{i}) \\
&\mathbb{E}\bm{n}_i= 0
   \end{array}
   \right\rbrace \quad i \in \mathbb{N}
\end{align}

Suppose to begin with that we are able to recover each $\bm{s} + \bm{n}_i$ \emph{without} the usual component-wise invertible functions. Then, writing $\bm{n}$ to denote all of the $\bm{n}_i$, it is possible to estimate $\bm{s}$ as
\begin{align*}
    \bm{s} \approx \Omega^N(\bm{s}, \bm{n}) = \frac{1}{N}\sum_{i=1}^N \left(\bm{s} + \bm{n}_i\right).
\end{align*}
Subject to mild conditions on the rate of growth of the variances $\text{Var}(\bm{n}_i)$ as $i\to\infty$, Kolmogorov's strong law implies that $\Omega^N(\bm{s}, \bm{n})$ is a good approximation to $\bm{s}$ as $N\to\infty$ in the sense that  $\Omega^N(\bm{s}, \bm{n}) \overset{a.s.}{\longrightarrow} \bm{s}$.
This implies moreover that it is possible to reconstruct the $\bm{n}_i$ by considering the residue $R^N_i(\bm{s}, \bm{n}) = (\bm{s} + \bm{n}_i) - \Omega^N(\bm{s}, \bm{n}) \overset{a.s.}{\longrightarrow} \bm{n}_i$.

In the presence of the unknown functions $\bm{k}_i$, we would be able to reconstruct $\bm{s}$ and the $\bm{n}_i$ if we were able to identify the inverses $\bm{e}_i = \bm{k}_i^{-1}$ for each $i$.
For any component-wise invertible functions $\bm{e}_i$, define
\begin{align*}
    \Omega_{\eb}^N(\bm{s}, \bm{n}) &= \frac{1}{N} \sum_{i=1}^N \bm{e}_i\circ \bm{k}_i( \bm{s} + \bm{n}_i) \\
    R_{\bm{e}, i}^N(\bm{s}, \bm{n}) &= \bm{e}_i\circ \bm{k}_i( \bm{s} + \bm{n}_i) - \Omega_{\bm{e}}^N(\bm{s}, \bm{n}).\\
\end{align*}
$\bm{e}_i$ is something we can choose and $\bm{k}_i(\bm{s}+\bm{n}_i) = \bm{h}_i(\bm{x}_i)$ is the output of the algorithm, and hence $\Omega_{\eb}^N(\bm{s}, \bm{n})$ and $R_{\bm{e}, i}^N(\bm{s}, \bm{n})$ are random variables with known distributions.
Subject to mild conditions, the dependence of these quantities on most or all of the $\bm{n}_i$ becomes increasingly small as $N$ grows and disappears in the limit $N\to\infty$.

\begin{lemma}\label{lem:last-lemma}
Suppose that the sequence $\mathbb{E}_{\bm{n}}[\Omega_{\eb}^N(\bm{s}, \bm{n})] = \frac{1}{N}\sum_{i=1}^N \mathbb{E}_{\bm{n}_i}[\bm{e}_i\circ \bm{k}_i( \bm{s} + \bm{n}_i)] $ converges as $N \to \infty$ for almost all $\bm{s}$, and write
\begin{align*}
    \Omega_\eb(\bm{s}) = \lim_{N\to\infty}\mathbb{E}_{\bm{n}}[\Omega_{\eb}^N(\bm{s}, \bm{n})].
\end{align*}

Suppose further that there exists $K$ such that $V_{\eb_i} = \mathrm{Var}\left(\bm{e}_i \circ \bm{g}_i(\bm{s} + \bm{n}_i) \right) \leq K$ for all $i$.
Then
\begin{align*}
    \Omega_{\eb}^N(\bm{s}, \bm{n}) & \overset{a.s.}{\longrightarrow} \Omega_{\eb}(\bm{s}) \\
    R_{\eb, i}^N(\bm{s}, \bm{n}) & \overset{a.s.}{\longrightarrow} R_{\eb, i}(\bm{s}, \bm{n}_i) = \bm{e}_i\circ \bm{k}_i( \bm{s} + \bm{n}_i) - \Omega_{\eb}(\bm{s})
\end{align*}
\end{lemma}

The proof can be found in Appendix \ref{appendix:last-lemma}.
Given some choice of $\bm{e}$, we can think of $\Omega_{\eb}(\bm{s})$ and $R_{\eb, i}(\bm{s}, \bm{n}_i)$ as our putative candidates for $\bm{s}$ and $\bm{n}_i$ respectively.
As discussed earlier, if we could identify $\bm{e}_i=\bm{k}_i^{-1}$, then we would have $\Omega_{\eb}(\bm{s}) = \bm{s}$ and $R_{\eb, i}(\bm{s}, \bm{n}_i) = \bm{n}_i$, and thus $\Omega_{\eb}$ and $R_{\eb, i}$ would satisfy the same independences and other statistical properties as $\bm{s}$ and $\bm{n}_i$ respectively.
Can we use these properties as criteria to identify good choices of $\bm{e}_i$?

The following theorem gives a set of sufficient conditions under which each $\bm{e}_i$ inverts $\bm{k}_i$ up to some affine ambiguity which is the same for every $i$.

\begin{theorem}
\label{thm:lastthm}
Suppose there exists $C>0$ such that $\text{Var}(\bm{n}_i) \leq C$ for all $i$ and let $\mathcal{G}_K = \big\lbrace
\{\bm{e}_i \}$ s.t.
\begin{align}
    & V_{\bm{e}_i} \leq K \ \forall i \label{eq:resid_1}\\
    & \Omega_{\bm{e}}(\bm{s}) < \infty \  \text{ for almost all } \bm{s} \label{eq:resid_2}\\
    &R_{\bm{e}, i} \independent R_{\bm{e}, j} \ \forall i \not= j, \label{eq:resid_3}\\
    \\    &\mathbb{E} R_{\bm{e}, i} = 0 \ \forall i \label{eq:resid_5} \\
    &R_{\bm{e}, i}(\bm{s}, \bm{n}_i) = R_{\bm{e}, i}(\bm{n}_i) \ \forall i \ \big\rbrace \label{eq:resid_6}
\end{align}

Then,
\begin{align*}
    \mathcal{G}_K \subseteq\left\lbrace \{ \bm{\alpha} \bm{k}^{-1}_i + \bm{\beta} \} \ : \ \bm{\alpha} \in \mathbb{R}^{D}_{\not=0}, \: \bm{\beta} \in \mathbb{R}^{D} \right\rbrace
\end{align*}
where $\bm\alpha \bm{k}^{-1}_i$ denotes the element-wise product with the scalar elements of $\bm{\alpha}$.
If $K \geq \text{Var}(\bm{s}) + C$, then $ \{ \bm{k}^{-1}_i \}  \in \mathcal{G}_K$,
and so $\mathcal{G}_K$ is non-empty for $K$ sufficiently large.
\end{theorem}
The proof can be found in Appendix \ref{sec:lasttmpr}.
It follows that it is possible recover $\bm{s}$ and $\bm{n}_i$ up to $\bm{\alpha}$ and $\bm{\beta}$ via $\Omega_\eb(\bm{s}) = \bm{\alpha}\bm{s} + \bm{\beta}$ and $R_{\bm{e}, i}(\bm{n}_i) = \bm{\alpha}\bm{n}_i$.

We remark that each of the conditions \ref{eq:resid_1}--\ref{eq:resid_5} can be verified from known information.
We conjecture that condition \ref{eq:resid_6} can be relaxed to assuming the verifiable condition of independence between $\Omega_{\eb}(\bm{s})$ and $R_{\eb, i}(\bm{s}, \bm{n}_i)$ for all $i$ along with additional regularity assumptions on the functional form of $R_{\eb, i}$ (e.g. smoothness).

To conclude, Theorem 8 provides sufficient conditions under which it is possible to fully reconstruct $\bm{s}$ with corrupted views.
In contrast to previous results in Sections \ref{sec:onenoisless} and \ref{sec:constrained}, this result leverages infinitely many corrupted views rather than vanishingly small corruption of finitely many views.

\section{RELATED WORK}
\label{sec:related-work}
A central concept in our work is that of multiple simultaneous views and joint extraction of features from them. We briefly review some related work considering similar settings.
\subsection{CANONICAL CORRELATION ANALYSIS}
\label{sec:probacca}
Given two (or more) random variables, the goal of Canonical Correlation Analysis (CCA)~\cite{hotelling1992relations} is to find a corresponding pair of linear subspaces that have high cross-correlation, so that each component within one of the subspaces is correlated with a single component from the other subspace~\cite{bishop2006pattern}.
In dealing with correlation instead of independence, CCA is more closely related to Principal Component Analysis (PCA) than to ICA.

CCA can be interpretated probabilistically~\cite{bach2005probabilistic} and is equivalent to maximum likelihood estimation in a graphical model which is a special case of that depicted in Figure \ref{fig:classic_hsr}.
The differences compared to our setting are (i) the latent components retrieved in CCA are forced to be uncorrelated, whereas our method is retrieves independent components; (ii) in CCA, mappings between the sources $\bm{s}$ and $\bm{x}$ are linear, whereas our method allows for nonlinear mappings.

At a high level, the model we consider in Section \ref{sec:constrained} is to CCA as nonlinear ICA is to PCA.
Nonlinear extensions of the basic CCA framework have been proposed~\cite{lai2000kernel, fukumizu2007statistical, andrew2013deep, michaeli2016nonparametric}, but identifiability results in the sense we consider in this paper are lacking.

\subsection{MULTI-VIEW LATENT VARIABLE MODELS}

Bearing a strong resemblance to our considered setting,~\cite{lederman2018learning} proposes a sequence of diffusion maps to find the common source of variability captured by multiple sensors, discarding irrelevant sensor-specific effects.
It computes the distance among the samples measured by different sensors to form a similarity matrix for the measurements of each sensor; each similarity matrix is then associated to a diffusion operator, which is a Markov matrix by construction. A Markov chain is then run by alternately applying these Markov matrices on the initial state. During these Markovian dynamics, sensor specific information will eventually vanish, and the final state will only contain information on the common source.
While the method focuses on recovering the common information in the form of a parametrization of the common variable, our method both inverts the mixing mechanisms of each view and recovers the common latent variables.

\cite{song2014nonparametric} proves identifiability for multi-view, latent variable models, unifying previously proposed spectral techniques~\cite{anandkumar2014tensor}. However, while the setting is similar to the one considered in this work, both the objectives and the employed methods are different.
The paper considers the setting in which $L$ variables $X_l$, $l=1, \ldots, L$ are observed; additionally, there exists an unobserved latent variable $H$, such that conditional distributions $P(X_l|H)$ are independent. While the setting bears obvious similarities with our multi-view ICA, the method proposed in~\cite{song2014nonparametric} is aimed at learning the mixture parameters, rather than the exact realization of latent variables.
Their method is based on the mean embedding of distributions in a Reproducing Kernel Hilbert Space and a result of identifiability for the parameters of the mean embeddings of $P(H)$ and $P(X|H)$ is proved.
Another related field of study is multi-view clustering, which considers a multiview setting and aims at performing clustering on a given dataset, see e.g.~\cite{de2005spectral} and~\cite{kumar2011co}. While related to our setting, this line of work is different from it in two key ways.
Firstly, clustering can be thought of as assigning a discrete latent label per datapoint. In contrast, our setting seeks to recover a continuous latent vector per datapoint.
Second, since no underlying generative model with discrete latent variable is assumed, identifiability results are not given.

\subsection{HALF-SIBLING REGRESSION}
\label{sec:hsr}
Half-sibling regression~\cite{scholkopf2016modeling} is a method to reconstruct a source from noisy observations by exploiting other sources that are affected by the same noise process but otherwise independent from it.

Suppose that a latent variable of interest $Q$ is not directly available, and that we can only observe corrupted versions of it, denoted as  $Y$, where the corruption is due to a noise $N$.
Without knowledge of $N$, it is impossible to reconstruct $Q$. However, if one or more additional variables $X$, also influenced by $N$, are observed, we can exploit them to model the effect of $N$ on $Y$ by regressing $Y$ on $X$.

Subtracting this from the observed $Y$ recovers the latent variable $Q$ up to a constant offset,
provided that (1) the additivity assumption
\[
Y = Q + f(N)
\]
holds, and (2) that $Y$ contains sufficient information about $f(N)$.
Analogous to our aim of recovering $\bm{s}$,
the goal of half-sibling regression is not to infer only the distribution of $Q$, but rather the random variable itself (almost surely).

\section{DISCUSSION AND CONCLUSION}
\label{sec:on_suffistv}
We presented identifiability results in a novel setting by extending the formalism of nonlinear ICA.
We have investigated different scenarios of multi-view latent variable models and provided theoretical proofs on the possibility of inverting the mixing function and recovering the sources in each case.
Our results thus extend the scarce literature on identifiability for nonlinear ICA models.

In the classical noiseless ICA setting, the deterministic relationship between the sources and observations means that inverting the mixing function and recovering the sources are equivalent.
In contrast, we consider views of corrupted versions of the common sources, resulting in the decoupling of the demixing and retrieval of the sources.
Remarkably, Theorem \ref{thm:lastthm} points towards the possibility of simultaneously solving the two problems in the limit of infinitely many views.

Classical nonlinear ICA is provably non-identifiable because a single view is not sufficiently informative to resolve non-trivial ambiguities when recovering the sources.
While many papers in the ICA literature have explored placing restrictions either on the source distribution or on the form of the mixing to resolve these ambiguities, in this paper we consider exploiting additional views to constrain the inverse problem.
Clearly, if a second view is identical to the first, then nothing is gained by its observation.
Hence, in order for the second view to assist in resolving ambiguity, it must be sufficiently different from the first.
This is the intuition behind the technical assumption of \emph{sufficiently distinct views}.

Typically, noise is a nuisance variable that would be preferably non-existent.
In our setting, however, the noise variables acting on the sources are a crucial component, without which the contrastive learning approach could not be applied.
Furthermore, the assumption of sufficiently distinct views is ultimately an assumption about the complexity of the joint distribution of the (corrupted) sources corresponding to each view.
Without the noise variables the sufficiently distinct views assumption could not hold.

Our setting is relevant in a number of practical real-world applications, namely in all datasets that include multiple distinct measurements of related phenomena.
In practice, it may be better to think of the noise variables rather as intrinsic sources of variability specific to each view.
In most practical applications this would probably not be a significant limitation due to the prevalence of stochasticity in real-world systems.

An exemplary application of our method can be found in the field of neuroimaging.
Consider a study involving a cohort of subjects (perceivers), measuring their response to the presentation of the same stimulus.
One of the key problems in the field is how to extract a shared response from all subjects despite high inter-subject variability and complex nonlinear mappings between latent source and observation~\cite{chen2015reduced, haxby2011common}.
Our results provide principled ways to extract and decompose the components of the shared response.
In particular, the setting described in our model is suited to account for the high variability of the responses throughout the cohort, since the measurement corresponding to each subject is given by a combination of individual variability and shared response.

Looking to the future, we note that Theorem \ref{thm:lastthm} builds on the setting of Theorem \ref{thm:two-noisy-views} which only makes use of pairwise information from the observations.
A natural extension of this work should investigate algorithms that explicitly make use of $N>2$ views, which we conjecture would allow relaxation of the additivity assumption on the corruptions.
Furthermore, Theorem \ref{thm:lastthm} provides results that only hold for the asymptotic limit as the number of views becomes large.
Other extensions to this result could include analysis of the case of finitely many views.

\subsubsection*{Acknowledgements}

Thanks to Krikamol Muandet for providing his office for fruitful discussions, to Matthias Bauer and Manuel W\"uthrich for proofreading and to Lucia Busso for interesting input about linguistics.

\bibliographystyle{plain}
\bibliography{main}

\clearpage

\clearpage
\appendix
\begin{center}
{\centering \LARGE APPENDIX}
\vspace{1cm}
\sloppy

\end{center}

\section{ON THE UNIDENTIFIABILITY OF NONLINEAR ICA}
\label{sec:on_unident}

The purpose of this section is to briefly review the proof of unidentifiablity of nonlinear ICA as~\cite{hyvarinen1999nonlinear}:
In this section we assume the most general conventional form of nonlinear ICA where the generative model follows:
\begin{equation}
    \label{eq:nonlinear_ica_noiseless}
    \xb=\fb(\bm{s})
\end{equation}
where $\bm{s}$ are the independent sources and $\xb$ are mixed signals. In the following, we show how to construct a function $\gb:R^n\to R^n$ so that the components $\yb=\gb(\xb)$ are independent. More importantly, we show that this construction is by no means unique.
\subsection{EXISTENCE}
The proposed method in~\cite{hyvarinen1999nonlinear} is a generalization of the famous Gram-Schmidt orthogonalization. Given $m$ independent variables, $y_1,\ldots,y_m$ and a variable $x$, one constructs a new variable $y_{m+1}=g(y_1,\ldots,y_m,x)$ so that the set $y_1s,\ldots,y_{m+1}$ is mutually independent. The construction process is defined recursively as follows. Assume we have $m$ independent random variables $y_1,\ldots,y_m$ with uniform distribution in $[0,1]^m$. $x$ is any random variable and $a_1,\ldots, a_m,b$ are some nonrandom scalars. Next, we define
\begin{align}
    g \left( a _ { 1 } , \ldots , a _ { m } , b ; p _ { y , x } \right) &= p ( x \leq b | y _ { 1 } = a _ { 1 } , \ldots , y _ { m } = a _ { m } )\nonumber\\
    &= \frac { \int _ { - \infty } ^ { b } p _ { y , x } \left( a _ { 1 } , \ldots , a _ { m } , \xi \right) \mathrm { d } \xi } { p _ { y } \left( a _ { 1 } , \ldots , a _ { m } \right) }
\end{align}
Theorem 1 of \cite{hyvarinen1999nonlinear} says that the random variable defined as $y_{m+1}=g(y_1,\ldots,y_m,x)$ is independent from the $y_1\ldots,y_m$ and $y_1,\ldots,y_{m+1}$ are uniformly distributed in the unit cube $[0,1]^{m+1}$.

\subsection{NON-UNIQUENESS}
In the previous section, it was shown that there exists a mapping $\gb$ that transforms any random vector $\xb$ into a uniformly distributed random vector $\yb=\gb(\xb)$. Here, we show that the construction of $\gb$ is not unique and this non-Uniqueness can be caused by several factors.
\begin{itemize}
    \item A linear transformation $\xb^\prime$ can precede the nonlinear map $\fb$ and then compute the independent components $\yb^\prime=\gb^\prime(\xb^\prime)$ where $\gb^\prime$ is computed as describe in the previous section. The new map $\gb^\prime$ gives a new decomposition of $\xb$ into independent components $\yb^\prime$ which can not be trivially reduced to $\yb$.
    \item An element-wise function $\hb$ can apply on the independent sources $\bm{s}$ first to give new sources $\bm{s}^\prime$ such that $s_i^\prime=h_i(s_i)$. Constructing the solution $g$ for these new scaled version of sources gives a new decomposition into independent components.
    \item Assume a class of measure-preserving automorphisms $\hb:[0,1]^n\to [0,1]^n$. The mapping $\hb$ does not change the probability distribution of a uniformly distributed random variable in $n$-dimensional hypercube. The composition $\hb\circ\gb$ gives another solution to nonlinear ICA. Therefore, the class of measure-preserving automorphisms gives a parameterization of the solutions to nonlinear ICA introducing a class of non-trivial indeterminacies.
\end{itemize}

\begin{figure}[t!]
    \centering
    \includegraphics[scale=0.3]{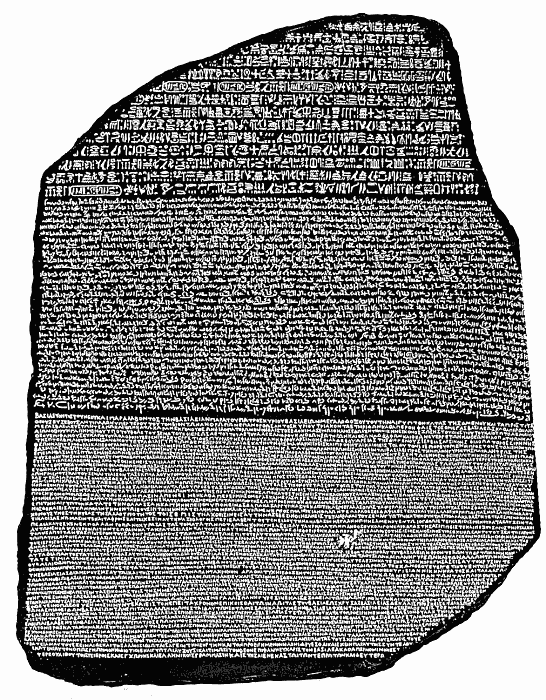}
    \caption{The Rosetta Stone, a stele found in 1799, inscribed with three versions of a decree issued at Memphis, Egypt in 196 BC. The top and middle texts are in Ancient Egyptian using hieroglyphic script and Demotic script, respectively, while the bottom is in Ancient Greek. (Source: Wikipedia)}
    \label{fig:rosetta}
\end{figure}
If only independence among the components matters, it is possible to construct a mapping $\bm{y}=G(\bm{x})$ such that $y_i$ is independent of $y_j$ for $i\neq j$ and uniformely distributed in $[0,1]^n$. This shows that at least one solution exists. The non-uniqueness of the solution can be shown by parameterising a class of infinitely many solutions. Once $\bm{y}$ is found with above conditions, any measure-preserving automorphism $f:[0,1]^n\to[0,1]^n$ can be used to parameterize $G$ as $f\circ G$, suggesting that there are infinitely many solutions to nonlinear ICA whose relations are nontrivial.
\subsection{THE SCALAR INVERTIBLE FUNCTION GAUGE}
\label{sec:gauge}
Another indeterminacy is element-wise functions $f_i$ applying on $y_i$ which suggets another dimension of ambiguity. Non-Gaussianity cannot help here since we can construct any marginal distribution by combining the CDF of the observed variable with the inverse CDF of the target marginal distribution. This indeterminacy is in some sense unavoidable and is related to the fact that in linear ICA recovery of the sources is possible up to a scalar multiplicative ambiguity.

\section{WHY DOES CLASSIFICATION RESULT IN THE LOG RATIO?}
\label{sec:converged}
Let us suppose that a variable $X$ is drawn with equal probability from two distributions $P_0$ and $P_1$ with densities $p_0(x)$ and $p_1(x)$ respectively.
We train a classifier $D: x \mapsto [0,1]$ to estimate the posterior probability that a particular realization of $X$ was drawn from $P_0$ with the cross entropy loss, i.e. the parameters of $D$ are chosen to minimize

\[
L(D) = \mathbb{E}_{X\sim P_0} \left[ - \log D(X) \right] + \mathbb{E}_{X\sim P_1} \left[ - \log (1 - D(X)) \right].
\]

As shown in, for instance, \cite{goodfellow2014generative}, the global optimum of this loss occurs when $D(x) = \frac{p_0(x)}{p_0(x) + p_1(x)}$, which can be rewritten as

\begin{align}
    D(x) &= \frac{1}{1 + p_1(x)/p_0(x)}\\
    &= \frac{1}{1 + \exp ( - \log (p_0(x)/p_1(x))) } \label{eq:density-ratio-classification}\\
\end{align}

Recall that in our setting, the function $r(x_1, x_2)$ is trained to classify between the two cases that $(x_1, x_2)$ is drawn from the joint distribution $\mathbb{P}_{x_1, x_2}$ (\emph{class $0$}) or the product of marginals $\mathbb{P}_{x_1}\mathbb{P}_{x_2}$ (\emph{class $1$}).
$r(x_1, x_2)$ is trained so that $\frac{1}{1 + \exp(-r(x_1, x_2))}$ estimates the posterior probability of $(x_1, x_2)$ belonging to class 0.
By comparing to Equation \ref{eq:density-ratio-classification}, it can be seen that

\begin{align*}
    r(x_1, x_2) &= \log \left( p(x_1, x_2) / p(x_1) p(x_2)\right) \\
    &= \log p(x_1 | x_2)  - \log p(x_1) \\
    &= \log p(x_2 | x_1)  - \log p(x_2) \\
\end{align*}

Note that in order for the classification trick of contrastive learning to be useful, the variables $x_1$ and $x_2$ cannot be deterministically related.
If this is the case, the log-ratio is everywhere either $0$ or $\infty$ and hence the learned features are not useful.

To see why this is the case, suppose that $x_1$, and $x_2$ are each $N$-dimensional vectors.
If they are deterministically related, $p(x_1, x_2)$ puts mass on an $N$-dimensional submanifold of a $2N$-dimensional space.
On the other hand, $p(x_1)p(x_2)$ will put mass on a $2N$-dim manifold since it is the product of two distributions each of which are N-dimensional.

In this case, the distributions $p(x_1, x_2)$ and $p(x_1)p(x_2)$ are therefore not absolutely continuous with respect to one another and thus the log-ratio is ill-defined: $p(x_1, x_2)/p(x_1)p(x_2) = \infty$ at any point $(x_1,x_2)$ at which $p(x_1, x_2)$ puts mass and zero at points where $p(x_1)p(x_2)$ puts mass and $p(x_1,x_2)$ does not.

\section{THE SUFFICIENTLY DISTINCT VIEWS ASSUMPTION}
\label{appendix:sdv}

We give the following two examples to provide intuition about the Sufficiently Distinct Views (SDV) assumption - one regarding a case in which it does not hold, and another one in which it does.

A simple case in which the assumption does not hold is when the conditional probability of $\bm{z}$ given $\bm{s}$ is Gaussian, as in

\begin{equation}
p(\bm{z}|\bm{s}) = \frac{1}{Z} \exp\left[ -\sum_i (z_i - s_i)^2/(2\sigma_i^2) \right]\,, \label{eq:unsatisfied}
\end{equation}

where $Z$ is the normalization factor, $Z = (2\pi)^{n/2}  \prod_i \sigma_i$.
Since taking second derivatives of the log-probability with respect to $s_i$ results in constants,
it can be easily shown that there is no way to find $2D$ vectors $\bm{z}_j$, $j=1, \ldots, 2D$, such that the corresponding $\bm{w}(\bm{s}, \bm{z}_j)$ (see Definition 1) are linearly independent.

The fact that the assumption breaks down in this case is reminiscent of the breakdown in the case of Gaussianity for linear ICA. Interestingly, in our work, the true latent sources \textbf{are} allowed to be Gaussian. In fact, the distribution of $\bm{s}$ does not enter the expression above.

An example in which the SDV assumption does hold is a conditional pdf given by

\begin{equation}
p(\bm{z}|\bm{s}) = \frac{1}{Z(\bm{s})} \exp \left[ - \sum_i (z_i^2  s_i^2 + z_i^4 s_i^4  ) \right]\,, \label{eq:satisfied}
\end{equation}

where $Z(\bm{s})$ is again a normalization function.
Proving that this distribution satisfies the SDV assumption requires a few lines of computation.
The idea is that $\bm{w}(\bm{s}, \bm{z})$ can be written as the product of a matrix and vector which are functions only of $\bm{s}$ and $\bm{z}$ respectively.
Once written in this form, it is straightforward to show that the columns of the matrix are linearly independent for almost all values of $\bm{s}$ and that $2D$ linearly independent vectors can be realized by different choices of $\bm{z}$.

\section{PROOF OF THEOREM \ref{thm:noiseless1} AND COROLLARY \ref{crl:noiseless1}}
\label{appendix:thm_noiseless}

\subsection{PROOF OF THEOREM \ref{thm:noiseless1}}\label{appendix:proof-thm1}
This proof is mainly inspired by the techniques employed by \cite{hyvarinen19a}.

\begin{proof}
We have to
show that, upon convergence, $h_{i}(\bm{x}_{1})$ are s.t.
\begin{align*}
h_{i}(\bm{x}_{1})\independent h_{j}(\bm{x}_{1}),\forall i\neq j
\end{align*}

We start by writing the difference in log-densities of the two classes:
\begin{align*}
\sum_{i}\psi_{i}(h_{i}(\bm{x}_{1}),\bm{x}_{2})  &=\sum_{i}\alpha_{i}(\bm{f}_{1,i}^{-1}(\bm{x}_{1}), \bm{f}_{2,i}^{-1}(\bm{x}_{2}))+\\
&-\sum_{i}\delta_{i}( \bm{f}_{2,i}^{-1}(\bm{x}_{2}))
\end{align*}
We now make the change of variables
\begin{align*}
\bm{y} & =\bm{h}(\bm{x}_{1})\\
\bm{v}(\bm{y}) & =\bm{f}_{1}^{-1}(\bm{h}^{-1}(\bm{y}))\\
\bm{t} & = \bm{f}_{2}^{-1}(\bm{x}_{2}))
\end{align*}
and rewrite the first equation in the following form:
\begin{align}
\sum_{i}\psi_{i}(y_{i},\bm{x}_{2})=&\sum_{i}\alpha_{i}(v_{i}(\bm{y}), t_{i})\\
-&\sum_{i}\delta_{i}( t_{i})
\end{align}

We take derivatives with respect to $y_j$, $y_{j'}$, $j \neq j'$,  of the LHS and RHS of equation \ref{eq:logistic}. Adopting the conventions in \ref{eq:convention1} and \ref{eq:convention2} and
\begin{align}
v^j_i(\bm{y})&=\partial v_i(\bm{y})/\partial y_j\\
v^{jj'}_i(\bm{y})&= \partial^2 v_i(\bm{y})/\partial y_j \partial y_{j'}\,,
\end{align}
we have
\begin{align*}
&\sum_{i} \alpha''_{i}(v_{i}(\bm{y}), t_{i})v^j_i(\bm{y})v^{j'}_i(\bm{y}) \\
&+ \alpha'_{i}(v_{i}(\bm{y}),  t_{i})v^{jj'}(\bm{y})=0\,,
\end{align*}
where taking derivative w.r.t. $y_j$ and $y_j'$ for $j \neq j'$ makes LHS equal to zero, since the LHS has functions which depend only one $y_i$ each.
If we now rearrange our variables by defining vectors $\bm{a}_i(\bm{y})$ collecting all entries $v_i^j(\bm{y})v_i^{j'}(\bm{y})$, $j=1, \ldots, n$, $j'=1, \ldots, j-1$, and vectors $\bm{b}_i(\bm{y})$ with the variables $v_i^j(\bm{y})v_i^{j'}(\bm{y})$, $j=1, \ldots, n$, $j'=1, \ldots, j-1$, the above equality can be rewritten as
\begin{align*}
&\sum_{i} \alpha''_{i}(v_{i}(\bm{y}), t_{i})\bm{a}_i(\bm{y}) \\
&+ \alpha'_{i}(v_{i}(\bm{y}),  t_{i}))\bm{b}_i(\bm{y})=0\,.
\end{align*}
The above expression can be recast in matrix form,
\[
\bm{M}(\bm{y})\bm{w}(\bm{y}, \bm{t})=0\,,
\]
where $\bm{M}(\bm{y}) = (\bm{a}_1(\bm{y}), \ldots,  \bm{a}_n(\bm{y}), \bm{b}_1(\bm{y}), \ldots, \bm{b}_n(\bm{y})) $ and $\bm{w}(\bm{y}, \bm{t}) = (\alpha''_{1}, \ldots, \alpha''_{n}, \alpha'_{1}, \ldots,\alpha'_{n})$. $\bm{M}(\bm{y})$ is therefore a $n(n-1)/2 \times 2n$ matrix, and $\bm{w}(\bm{y}, \bm{t})$ is a $2n$ dimensional vector.

To show that $\bm{M}(\bm{y})$ is equal to zero, we invoke the SDV assumption.
This implies the existence of $2n$ linearly independent $\bm{w}(\bm{y}, \bm{t}_j)$.
It follows that

\[
\bm{M}(\bm{y})[\bm{w}(\bm{y}, \bm{t}_1), \ldots, \bm{w}(\bm{y}, \bm{t}_{2n})]=0\,,
\]

and hence $\bm{M}(\bm{y})$ is zero by elementary linear algebraic results.
It follows that $v_i^j(\bm{y})\not=0$ for at most one value of $j$, since otherwise the product of two non-zero terms would appear in one of the entries of $\bm{M}(\bm{y})$, thus rendering it non-zero.
Thus $v_i$ is a function only of one $y_j$.

Observe that $\bm{v}(\bm{y}) = \bm{s}$.
We have just proven that $v_i(y_{\pi(i)}) = s_i$.
Since $v_i$ is invertible, it follows that $h_{\pi(i)}(\bm{x}_{1}) = y_{\pi(i)} = v_i^{-1}(s_i)$ and hence the components of $\bm{h}(\bm{x}_{1})$ recover the components of $\bm{s}$ up to the invertible component-wise ambiguity given by $\bm{v}$, and the permutation ambiguity.

\end{proof}

\subsection{PROOF OF COROLLARY \ref{crl:noiseless1}}\label{appendix:proof-cor2}
\begin{proof}
This follows exactly by repeating the proof of Theorem \ref{thm:noiseless1} where the roles of $\bm{x}_1$ and $\bm{x}_2$ are exchanged and the regression function in the statement of the corollary is used.
\end{proof}

\section{PROOF OF THEOREMS \ref{thm:demixing} AND \ref{thm:two-noisy-views}}
\label{appendix:thm1}

Theorem \ref{thm:demixing} is a special case of Theorem \ref{thm:two-noisy-views} by considering the case $\bm{g}_1(\bm{s}, \bm{n}_1) = \bm{s}$.
We therefore prove only the more general Theorem \ref{thm:two-noisy-views}.

\begin{proof}
We have to
show that, upon convergence, $h_{i}(\bm{x}_{1})$ and $k_{i}(\bm{x}_{2})$
are such that
\begin{align}
h_{1,i}(\bm{x}_{1})\independent h_{1,j}(\bm{x}_{1}),\forall i\neq j \label{eq:inds_twov_1} \\
h_{2,i}(\bm{x}_{2})\independent h_{2,j}(\bm{x}_{2}),\forall i\neq j \label{eq:inds_twov_2}\\
h_{1,i}(\bm{x}_{1})\independent h_{2,j}(\bm{x}_{2}),\forall i\neq j. \label{eq:inds_twov_3}
    \end{align}

We start by exploiting Equations \ref{eq:noisylogdens_1} and \ref{eq:noisylogdens_2} to write the difference in log-densities of the two classes
\begin{align}
&\sum_{i}\psi_{i}(h_{1,i}(\bm{x}_{1}),h_{2,i}(\bm{x}_{2}))\nonumber\\
=&\sum_{i}\eta_{i}(\bm{f}_{1,i}^{-1}(\bm{x}_{1}), \bm{f}_{2,i}^{-1}(\bm{x}_{2})) - \sum_{i}\theta_{i}(\bm{f}_{1,i}^{-1}(\bm{x}_{1})) \label{eq:first_factorization}\\
=&\sum_{i}\lambda_{i}(\bm{f}_{2,i}^{-1}(\bm{x}_{2}), \bm{f}_{1,i}^{-1}(\bm{x}_{1})) - \sum_{i}\mu_{i}(\bm{f}_{2,i}^{-1}(\bm{x}_{2}))\label{eq:2nd_factorization}
\end{align}
We now make the change of variables
\begin{align*}
\bm{y} & =\bm{h}_1(\bm{x}_{1})\\
\bm{t} & =\bm{h}_2(\bm{x}_{2})\\
\bm{v}(\bm{y}) & =\bm{f}_{1}^{-1}(\bm{h}_1^{-1}(\bm{y}))\\
\bm{u}(\bm{t}) & =\bm{f}_{2}^{-1}(\bm{h}_2^{-1}(\bm{t}))
\end{align*}
and rewrite equation \ref{eq:first_factorization} in the following form:
\begin{align}
&\sum_{i}\psi_{i}(y_{i},t_{i}) \nonumber \\
&=\sum_{i}\eta_{i}(v_i(\bm{y}), u_i(\bm{t}))
-\sum_{i}\theta_{i}(v_i(\bm{y}))\label{eq:logistic}
\end{align}
We first want to prove the condition in Equation \ref{eq:inds_twov_1}.
We will show this is true by proving that
\begin{equation}
\label{eq:v_onev}
v_{i}(\bm{y})  \equiv v_{i}(y_{\pi(i)})\\
\end{equation}
for some permutation of the indices $\pi$ with respect to the indexing of the sources $\bm{s} = (s_1, \ldots, s_D)$.

We take derivatives with respect to $y_j$, $y_{j'}$, $j \neq j'$,  of the LHS and RHS of equation \ref{eq:logistic}, yielding
\begin{align*}
&\sum_{i} \eta''_{i}(v_{i}(\bm{y}),u_{i}(\bm{t}))v^j_i(\bm{y})v^{j'}_i(\bm{y}) \\
&\ + \sum_{i}\eta'_{i}(v_{i}(\bm{y}), u_{i}(\bm{t}))v^{jj'}(\bm{y})=0
\end{align*}
If we now rearrange our variables by defining vectors $\bm{a}_i(\bm{y})$ collecting all entries $v_i^j(\bm{y})v_i^{j'}(\bm{y})$, $j=1, \ldots, n$, $j'=1, \ldots, j-1$, and vectors $\bm{b}_i(\bm{y})$ with the variables $v_i^j(\bm{y})v_i^{j'}(\bm{y})$, $j=1, \ldots, n$, $j'=1, \ldots, j-1$, the above equality can be rewritten as

\begin{align*}
&\sum_{i} \eta''_{i}(v_{i}(\bm{y}),u_{i}(\bm{t}))\bm{a}_i(\bm{y}) \\
&+ \eta'_{i}(v_{i}(\bm{y}), u_{i}(\bm{t}))\bm{b}_i(\bm{y})=0\,.
\end{align*}

Again following \cite{hyvarinen19a}, we recast the above formula in matrix form,

\begin{equation}
\label{eq:matrixmult}
\bm{M}(\bm{y})\bm{w}(\bm{y}, \bm{t})=0\,,
\end{equation}

where $\bm{M}(\bm{y}) = (\bm{a}_1(\bm{y}), \ldots,  \bm{a}_n(\bm{y}), \bm{b}_1(\bm{y}), \ldots, \bm{b}_n(\bm{y})) $ and $\bm{w}(\bm{y}, \bm{t}) = (\eta''_{1}, \ldots, \eta''_{n}, \eta'_{1}, \ldots,\eta'_{n})$. $\bm{M}(\bm{y})$ is therefore a $n(n-1)/2 \times 2n$ matrix, and $\bm{w}(\bm{y}, \bm{t})$ is a $2n$ dimensional vector.

To show that $\bm{M}(\bm{y})$ is equal to zero, we invoke the SDV assumption on $\bm{\eta}$.
This implies the existence of $2n$ linearly independent $\bm{w}(\bm{y}, \bm{t}_j)$.
It follows that

\[
\bm{M}(\bm{y})[\bm{w}(\bm{y}, \bm{t}_1), \ldots, \bm{w}(\bm{y}, \bm{t}_{2n})]=0\,,
\]

and hence $\bm{M}(\bm{y})$ is zero by elementary linear algebraic results.
It follows that $v_i^j(\bm{y})\not=0$ for at most one value of $j$, since otherwise the product of two non-zero terms would appear in one of the entries of $\bm{M}(\bm{y})$, thus rendering it non-zero.
Thus $v_i$ is a function only of one $y_j = y_{\pi(i)}$.

Observe that $\bm{v}(\bm{y}) = \bm{s}$.
We have just proven that $v_i(y_{\pi(i)}) = s_i$.
Since $v_i$ is invertible, it follows that $h_{\pi(i)}(\bm{x}_{1}) = y_{\pi(i)} = v_i^{-1}(s_i)$ and hence the components of $\bm{h}(\bm{x}_{1})$ recover the components of $\bm{s}$ up to the invertible component-wise ambiguity given by $\bm{v}$, and the permutation ambiguity.

For the condition in Equation \ref{eq:inds_twov_2}, we need
\begin{equation}
\label{eq:u_onev}
u_{i}(\bm{t})  \equiv u_{i}(t_{\tilde{\pi}(i)})\,,
\end{equation}
where the permutation $\tilde{\pi}$ doesn't need to be equal to $\pi$.
By symmetry, exactly the same argument as used to prove the condition in Equation \ref{eq:v_onev} holds, by replacing $(\bm{v},\bm{y}, \bm{\eta}, \bm{\theta})$ with $(\bm{u},\bm{t}, \bm{\lambda}, \bm{\mu})$, noting that the SDV assumption is also assumed for $\bm{\lambda}$.
\\
We have shown that $\bm{y}=\bm{h}_1(\bm{x}_1)$ and $\bm{t}=\bm{h}_2(\bm{x}_2)$ estimate $\bm{g}_1(\bm{s}, \bm{n}_1)$ and $\bm{g}_2(\bm{s}, \bm{n}_2)$ up to two different gauges of all possible scalar invertible functions.

A remaining ambiguity could be that the two representations might be misaligned; that is, defining $\bm{z}_1=\bm{g}_1(\bm{s}, \bm{n}_1)$ and $\bm{z}_2=\bm{g}_2(\bm{s}, \bm{n}_2)$, while
\begin{equation}
z_{1,i} \independent z_{2,j} \forall i \neq j \label{eq:fact}
\end{equation}
we might have
\[
y_{\pi(i)} \independent t_{\tilde{\pi}(j)} \forall i \neq j\,,
\]
where $\pi(i)$, $\tilde{\pi}(i)$ are two different permutations of the indices $i=1, \ldots, n$. We want to show that this ambiguity is also resolved; that means, our goal is to show that
\begin{equation}
y_{i} \independent t_{j},\;\;\forall i \neq j \label{eq:aim_lastpart}
\end{equation}

We recall that, by definition, we have $v_i(y_{\pi(i)}) = z_{1,i}$ and $u_j(t_{\tilde{\pi}(j)}) = z_{2,j}$. Then, due to equation \ref{eq:fact},
\begin{align}
v_i(y_{\pi(i)}) & \independent u_j(t_{\tilde{\pi}(j)}) \,\,\, \forall i \neq j \label{eq:permutind_1}\\
\implies y_{\pi(i)} & \independent t_{\tilde{\pi}(j)} \,\,\, \forall i \neq j \label{eq:permutindep}\\
\implies y_{i} & \independent t_{\tilde{\pi}\circ \pi^{-1} (j)} \,\,\, \forall i \neq j\,, \label{eq:permutind_2}
\end{align}
where the implication \ref{eq:permutind_1}-\ref{eq:permutindep} follows from invertibility of $v_i$ and $u_j$, and the implication \ref{eq:permutindep}-\ref{eq:permutind_2} follows from considering that, given that we know \ref{eq:permutindep}, we can define $l=\pi(j)$ and $k=\pi(i)$ and have
\[
y_{k}  \independent t_{\tilde{\pi} \circ \pi^{-1} (l)} \,\,\, \forall k \neq l.
\]

Define
\[
\tau = \tilde{\pi} \circ \pi^{-1}
\]
and note that it is a permutation. Then
\begin{equation}
    y_i \independent t_{\tau(j)}  \forall i \neq j \label{eq:tauperm}
\end{equation}

Fix any particular $i$.
Our goal is to show that for any $j\not= i$ the independence relation in Equation \ref{eq:aim_lastpart} holds.
There are two possibilities:
\begin{enumerate}[i]
\item $\tau(i)=i$
\item $\tau(i)\neq i$
\end{enumerate}
In the first case, $\tau$ restricted to the set $\{1,\ldots,D\}\setminus\{i\}$ is still a permutation, and thus considering the independences of Equation \ref{eq:tauperm} for all $j\not= i$ implies each of the independences of Equation \ref{eq:aim_lastpart} and we are done.

Let us consider the second case. Then,
\[
\exists l \in \{1, \ldots, D \}\setminus\{i\}\,\, \text{s.t.} \,\, l = \tau(i)\,.
\]
We then need to prove
\begin{equation}
y_i \independent t_l\,, \label{eq:ref_indices}
\end{equation}

which is the only independence implied by Equation \ref{eq:aim_lastpart} which is not implied by Equation \ref{eq:tauperm}.

In order to do so, we rewrite equation \ref{eq:logistic}, yielding
\begin{align}
&\sum_{m}\psi_{m}(y_{m},t_{m}) \nonumber \\
=&\sum_{m}\eta_{m}(v_m(y_{\pi(m)}), u_m(t_{\tilde{\pi}(m)}))
-\sum_{m}\theta_{i}(v_m(y_{\pi(m)}))
\end{align}
We now take derivative with respect to $y_i$ and $t_l$ in \ref{eq:ref_indices}; noting that $\tilde{\pi}^{-1}(l) = \pi^{-1}(i) $, we get
\begin{align}
    0 = & \frac{\partial^2}{\partial v_{\pi^{-1}(i)} \partial u_{\pi^{-1}(i)}} \eta_{\pi^{-1}(i)}(v_{\pi^{-1}(i)}(y_i), u_{\pi^{-1}(i)}(t_l)) \nonumber \\
    \times & \frac{\partial}{\partial y_i}v_{\pi^{-1}(i)}(y_i) \frac{\partial }{\partial t_l} u_{\pi^{-1}(i)}(t_l) \label{eq:perm_deriv}
\end{align}

Since $v_{\pi^{-1}(i)}(y_i)$ is a smooth and invertible function of its argument, the set of $y_i$ such that $\frac{\partial}{\partial y_i}v_{\pi^{-1}(i)}(y_i) = 0$ has measure zero.
Similarly, $\frac{\partial }{\partial t_l} u_{\pi^{-1}(i)}(t_l) = 0$ on a set of measure zero.

It therefore follows that
\begin{align*}
    \frac{\partial}{\partial y_i}v_{\pi^{-1}(i)}(y_i) \frac{\partial }{\partial t_l} u_{\pi^{-1}(i)}(t_l) \neq 0
\end{align*}
almost everywhere and hence that
\begin{equation}
\frac{\partial^2}{\partial v_{\pi^{-1}(i)} \partial u_{\pi^{-1}(i)}} \eta_{\pi^{-1}(i)}(v_{\pi^{-1}(i)}(y_i), u_{\pi^{-1}(i)}(t_l)) = 0\,. \label{eq:additive_eta}
\end{equation}
almost everywhere.
We can thus conclude that
\begin{align*}
&\eta_{\pi^{-1}(i)}(v_{\pi^{-1}(i)}(y_i), u_{\pi^{-1}(i)}(t_l)) = \\ &\eta_{\pi^{-1}(i)}^y(v_{\pi^{-1}(i)}(y_i))+ \eta_{\pi^{-1}(i)}^t(u_{\pi^{-1}(i)}(t_l))
\end{align*}
This in turn implies that, for some functions $A$ and $B$, we can write
\begin{align*}
    &\log p(z_{1, \pi^{-1}(i)}|z_{2, \pi^{-1}(i)}) - \log p(z_{1, \pi^{-1}(i)}) \\ &= A(v_{\pi^{-1}(i)}(y_i)) + B(u_{\pi^{-1}(i)}(t_l))
\end{align*}
and therefore
\begin{align*}
    \log p(z_{1, \pi^{-1}(i)},z_{2, \pi^{-1}(i)}) = C(v_{\pi^{-1}(i)}(y_i)) + D(u_{\pi^{-1}(i)}(t_l))
\end{align*}
for some functions $C$ and $D$. This decomposition of the log-pdf implies
\begin{align*}
z_{1, \pi^{-1}(i)} &\independent z_{2, \pi^{-1}(i)}\\
\implies z_{1, \pi^{-1}(i)} &\independent z_{2, \tilde{\pi}^{-1}(l)}  \\
\implies v_{\pi^{-1}(i)}(y_i)  &\independent u_{\tilde{\pi}^{-1}(l)}(t_l) \\
\implies y_i  &\independent t_l \,,
\end{align*}
where the last implication holds due to invertibility of $v_{\pi^{-1}(i)}$ and $u_{\tilde{\pi}^{-1}(l)}$.

We have thus concluded the proof.

\end{proof}

\section{PROOF OF COROLLARY \ref{crl:lownoise}}
\label{appendix:thm2}

\begin{proof}
Denoting by $\bm{d}^{(k)}_1$ the component-wise invertible ambiguity up to which $\bm{g}(\bm{s}, \bm{n}_1^{(k)})$ is recovered, we have that
\begin{align}
     &\inf_{\bm{e}\in \bm{E}} \mathbb{E}_{\bm{x}_1} \left[ \left \|\bm{s} - \bm{e}(\bm{h}_1^{(k)}(\bm{x}_1)) \right \|_2^2 \right]\\
    &=\inf_{\bm{e}\in \bm{E}} \mathbb{E}_{(\bm{n}_1^{(k)}, \bm{s})} \left[ \left \|\bm{s} - \bm{e} \circ \bm{d}^{(k)}_1 \circ \bm{g}_1(\bm{s}, \bm{n}_1^{(k)}) \right \|_2^2 \right]\\
    &=\inf_{\tilde{\bm{e}}\in \bm{E}} \mathbb{E}_{(\bm{n}_1^{(k)}, \bm{s})} \left[ \left \|\bm{s} - \tilde{\bm{e}} \circ  \bm{g}_1(\bm{s}, \bm{n}_1^{(k)}) \right \|_2^2 \right]  \\
    &\leq\mathbb{E}_{(\bm{n}_1^{(k)}, \bm{s})} \left[ \left \|\bm{s} - \bm{e^*} \circ  \bm{g}_1(\bm{s}, \bm{n}_1^{(k)}) \right \|_2^2 \right]\label{eq:low_bounded}
\end{align}
The lower bound holds for any $\bm{e^*}\in\bm{E}$ by definition of infimum and in particular for $\bm{e^*} = \bm{g}_1 |^{-1}_{\bm{n}=0}$, the existence of which is guaranteed by the assumptions on $\bm{g}_1$.
Taking a Taylor expansion of $\bm{e^*} \circ  \bm{g}_1(\bm{s}, \bm{n}_1^{(k)})$ around $\bm{n}_1^{(k)}=0$ yields
\begin{align*}
     &\mathbb{E}_{(\bm{n}_1^{(k)}, \bm{s})} \Bigg[  \Bigg\|\bm{s} - \bm{e^*} \circ \bm{g}_1 (\bm{s}, 0) \\
     &\quad+ \left.\left.\frac{\partial \bm{e^*}}{\partial \bm{g}_1} \frac{\partial \bm{g}_1 (\bm{s}, 0)}{\partial \bm{n}_1^{(k)}} \cdot \bm{n}_1^{(k)} + \mathcal{O}(\|\bm{n}_1^{(k)}\|^2) \right \|_2^2 \right]\\
     &=\mathbb{E}_{(\bm{n}_1^{(k)}, \bm{s})} \left[ \left \|\frac{\partial \bm{e^*}}{\partial \bm{g}_1} \frac{\partial \bm{g}_1 (\bm{s}, 0)}{\partial \bm{n}_1^{(k)}} \cdot \bm{n}_1^{(k)} + \mathcal{O}(\|\bm{n}_1^{(k)}\|^2) \right \|_2^2 \right]\\
     &\longrightarrow 0 \text{ as $k \longrightarrow \infty$}
\end{align*}
where the last equality follows from fact that $\bm{e^*} = \bm{g} |^{-1}_{\bm{n}=0}$ and the convergence follows from the fact that $\bm{n}_1^{(k)} \longrightarrow 0$ as $k \to \infty$.
\end{proof}

\section{PROOF OF LEMMA \ref{lem:last-lemma}}
\label{appendix:last-lemma}

We will make crucial use of \emph{Kolmogorov's strong law}:
\begin{theorem}
Suppose that $X_n$ is a sequence of independent (but not necessarily identically distributed) random variables with
\begin{align*}
    \sum_{n=1}^\infty \frac{1}{n^2}\mathrm{Var} [X_n] < \infty
\end{align*}
Then,
\begin{align*}
    \frac{1}{N}\sum_{n=1}^N X_n - \mathbb{E}[X_n] \overset{a.s.}{\longrightarrow} 0
\end{align*}
\end{theorem}

Fix $\bm{s}$ and consider $\Omega_{\eb}^N(\bm{s}, \bm{n})$ as a random variable with randomness induced by $\bm{n}$.
We will show that for almost all $\bm{s}$ this converges $\bm{n}$-almost surely to a constant, and hence $\Omega_{\eb}^N(\bm{s}, \bm{n})$ converges almost surely to a function of $\bm{s}$.

The law of total expectation says that
\begin{align*}
    &\mathrm{Var}_{\bm{s}, \bm{n}_i} [\bm{e}_i\circ \bm{k}_i(\bm{s} + \bm{n}_i)] \\
    &= \mathbb{E}_{\bm{s}}\left[ V_i(\bm{s}) \right] + \mathrm{Var}_{\bm{s}}\left[ \mathbb{E}_{\bm{n}_i} [\bm{e}_i\circ \bm{k}_i( \bm{s} + \bm{n}_i)] \right] \\
    & \geq \mathbb{E}_{\bm{s}}\left[ V_i(\bm{s}) \right].
\end{align*}
Since by assumption $\mathrm{Var}_{\bm{s}, \bm{n}_i} [\bm{e}_i\circ \bm{k}_i(\bm{s} + \bm{n}_i)] \leq K$, we have that
\begin{align*}
    \mathbb{E}_{\bm{s}}\left[ \sum_{i=1}^\infty  \frac{V_i(\bm{s})}{i^2} \right] \leq \frac{ K \pi^2}{6}
\end{align*}
and therefore  $\sum_{i=1}^\infty  \frac{V_i(\bm{s})}{i^2} < \infty$ with probability $1$ over $\bm{s}$, else the expectation above would be unbounded since $V_i(\bm{s})\geq 0$.

We have further that for almost all $\bm{s}$,
\begin{align*}
    \Omega_{\bm{e}}(\bm{s}) = \lim_{N\to\infty}\frac{1}{N}\sum_{i=1}^N E_{\bm{e}_i}(\bm{s})
\end{align*}
exists.
Therefore, for almost all $s$ the conditions of Kolmogorov's strong law are met by $\Omega_{\bm{e}}^N(\bm{s}, \bm{n})$ and so
\begin{align*}
    \Omega_{\bm{e}}^N(\bm{s}, \bm{n}) - \mathbb{E}_{\bm{n}}[\Omega_{\bm{e}}^N(\bm{s}, \bm{n})] \overset{\bm{n}-a.s.}{\longrightarrow} 0
\end{align*}

Since $\mathbb{E}_{\bm{n}}[\Omega_{\bm{e}}^N(\bm{s}, \bm{n})] \overset{\bm{n}-a.s.}{\longrightarrow} \Omega_{\bm{e}}(\bm{s})$, it follows that
\begin{align*}
    \Omega_{\bm{e}}^N(\bm{s}, \bm{n}) \overset{\bm{n}-a.s.}{\longrightarrow} \Omega_{\bm{e}}(\bm{s}).
\end{align*}
Since this holds with probability $1$ over $\bm{s}$, we have that
\begin{align*}
    \Omega_{\bm{e}}^N(\bm{s}, \bm{n}) \overset{\bm{n}-a.s.}{\longrightarrow} \Omega_{\bm{e}}(\bm{s}).
\end{align*}

It follows that we can write

\begin{align*}
    R_{\bm{e}, i}^N(\bm{s}, \bm{n}) &= \bm{e}_i\circ \bm{k}_i( \bm{s} + \bm{n}_i) - \Omega_{\bm{e}}^N(\bm{s}, \bm{n}) \\
    &\overset{a.s.}{\longrightarrow} R_{\bm{e}, i}(\bm{s}, \bm{n}_i):= \bm{e}_i\circ \bm{k}_i( \bm{s} + \bm{n}_i) - \Omega_{\bm{e}}(\bm{s})
\end{align*}

\section{PROOF OF THEOREM \ref{thm:lastthm}}
\label{sec:lasttmpr}

We will begin by showing that if $K \geq \mathrm{Var}(\bm{s}) + C$ then $\{ \bm{k}^{-1}_i \}  \in \mathcal{G}_K$.

For $\bm{e}_i = \bm{k}_i^{-1}$, we have that
\begin{align*}
    \Omega_{\bm{e}}^N(\bm{s}, \bm{n}) = \frac{1}{N} \sum_{i=1}^N \bm{s} + \bm{n}_i &\overset{a.s.}{\longrightarrow} \bm{s} = \Omega_{\bm{e}}^N(\bm{s})\\
    R_i^N = \bm{s} + \bm{n}_i - \Omega_{\bm{e}}(\bm{s}, \bm{n})  &\overset{a.s.}{\longrightarrow} \bm{n}_i = R_{\bm{e}, i}(\bm{n}_i)
\end{align*}
where the convergences follow from application of Kolmogorov's strong law, using the fact that $\mathrm{Var}(\bm{n}_i) \leq C$ for all $i$.
Satisfaction of condition \ref{eq:resid_1} follows from the fact that $\mathrm{Var}_{\bm{s}, \bm{n}_i} (\bm{s} + \bm{n}_i) \leq C + \mathrm{Var}(\bm{s}) \leq K$.
Since $\bm{s}$ is a well-defined random variable,  $\Omega_{\bm{e}}(\bm{s}) < \infty$ with probability $1$, satisfying condition \ref{eq:resid_2}.
It follows from the mutual independence of $\bm{n}_i$ and $\bm{n}_j$ that $R_{\bm{e}, i}$ and $R_{\bm{e}, j}$ satisfy condition \ref{eq:resid_3}.
Condition \ref{eq:resid_5} follows from the fact that $\mathbb{E}[\bm{n}_i]=0$
Condition \ref{eq:resid_6} follows from $R_{\bm{e}, i}$ being constant as a function of $\bm{s}$.

It therefore follows that $\{ \bm{k}^{-1}_i \}  \in \mathcal{G}_K$ for $K$ sufficiently large.

We will next show that if $\{ \bm{e}_i\} \in \mathcal{G}_K$ then there exist a matrix $\bm{\alpha}$ and vector $\bm{\beta}$ such that $\bm{e}_i = \bm{\alpha} \bm{k}_i^{-1} + \bm{\beta}$ for all $i$.
Since $\bm{e}_i$ acts coordinate-wise, it moreover follows that $\bm{\alpha}$ is diagonal.

First, we will show that each $\bm{e}_i\circ\bm{k}_i$ is affine, i.e. there exist potentially different $\bm{\alpha}_i, \bm{\beta}_i$ such that $\bm{e}_i = \bm{\alpha}_i \bm{k}_i^{-1} + \bm{\beta}_i$ for each $i$.

Then we will show that we must have $\bm{\alpha}_i = \bm{\alpha}_j$ and $\bm{\beta}_i = \bm{\beta}_j$ for all $i,j$.

To see that $\bm{e}_i$ is affine, we make use of that fact that $R_{\bm{e},i}$ is constant as a function of $\bm{s}$.
It follows that for any $x$ and $y$
\begin{align*}
    \bm{e}_i\circ\bm{k}_i(x + y) &= R_{\bm{e},i}(x) + \Omega_{\bm{e}}(y) \\
    &= R_{\bm{e},i}(x) + \Omega_{\bm{e}}(0) + R_{\bm{e},i}(0) + \Omega_{\bm{e}}(y) \\
    & \qquad- \left(R_{\bm{e},i}(0) +  \Omega_{\bm{e}}(0)\right) \\
    &= \bm{e}_i\circ\bm{k}_i(x) + \bm{e}_i\circ\bm{k}_i(y) - \bm{e}_i\circ\bm{k}_i(0)
\end{align*}
It therefore follows that $\bm{e}_i\circ\bm{k}_i$ is affine, since if we define
\begin{align*}
    L(x + y) &= \bm{e}_i\circ\bm{k}_i(x + y) - \bm{e}_i\circ\bm{k}_i(0) \\
    &= \left(\bm{e}_i\circ\bm{k}_i(x) - \bm{e}_i\circ\bm{k}_i(0)\right) \\
    & \qquad+ \left(\bm{e}_i\circ\bm{k}_i(y) - \bm{e}_i\circ\bm{k}_i(0)\right) \\
    &= L(x) + L(y)
\end{align*}
then $L$ is linear and we can write $\bm{e}_i\circ\bm{k}_i(x)$ as the sum of a linear function and a constant:
\begin{align*}
    \bm{e}_i\circ\bm{k}_i(x) = L(x) + \bm{e}_i\circ\bm{k}_i(0)
\end{align*}
Thus $\bm{e}_i\circ\bm{k}_i$ is affine, and we have some (diagonal) matrix $\bm{\alpha}_i$ and vector $\bm{\beta}_i$ such that for any $x$
\begin{align*}
    &\bm{e}_i\circ\bm{k}_i(x) = \bm{\alpha}_i x  + \bm{\beta}_i \\
    \implies& \bm{e}_i \left(x \right) = \bm{\alpha}_i \bm{k}_i^{-1} x + \bm{\beta}_i.
\end{align*}

Next we show that for the set of $\{\bm{e}_i = \bm{\alpha}_i \bm{k}_i^{-1} + \bm{\beta}_i\}$, it must be the case that each $\bm{\alpha}_i = \bm{\alpha}_j$ and $\bm{\beta}_i = \bm{\beta}_j$.

Observe that
\begin{align*}
    \Omega_{\bm{e}}^N(\bm{s}, \bm{n}) &= \frac{1}{N} \sum_{i=1}^N \bm{\alpha}_i \bm{s} + \bm{\alpha}_i \bm{n}_i + \bm{\beta}_i \\
    &= \left( \frac{1}{N} \sum_{i=1}^N \bm{\alpha}_i \right)\bm{s} + \frac{1}{N} \sum_{i=1}^N \bm{\beta}_i  + \frac{1}{N} \sum_{i=1}^N \bm{\alpha}_i \bm{\bm{n}}_i \\
    \mathbb{E}_{\bm{n}}[\Omega_{\bm{e}}^N(\bm{s}, \bm{n})] &= \left( \frac{1}{N} \sum_{i=1}^N \bm{\alpha}_i \right)\bm{s} +  \frac{1}{N} \sum_{i=1}^N \bm{\beta}_i
\end{align*}

Define
\begin{align*}
\bm{\alpha} &= \lim_{N\to\infty}\frac{1}{N}\sum_{i=1}^N \bm{\alpha}_i \\
\bm{\beta} &= \lim_{N\to\infty}\frac{1}{N}\sum_{i=1}^N \bm{\beta}_i \\
\end{align*}
which exist by the assumption that $\Omega_{\bm{e}}^N(\bm{s}, \bm{n})$ converges as $N\to\infty$.
Thus
\begin{align*}
    \Omega_{\bm{e}}(\bm{s}) &= \bm{\alpha} \bm{s} + \bm{\beta} \\
    R_{\bm{e}, i}(\bm{s}, \bm{n}_i) &= (\bm{\alpha}_i - \bm{\alpha})\bm{s} + \bm{\alpha}_i\bm{n}_i + \bm{\beta}_i - \bm{\beta}
\end{align*}
Now, suppose that there exist $i$ and $j$ such that such that $\bm{\alpha}_i \not= \bm{\alpha}_j$.
It follows that
\begin{align*}
    R_{\bm{e}, i}(\bm{s}, \bm{n}_i) &= (\bm{\alpha}_i - \bm{\alpha})\bm{s} + \bm{\alpha}_i\bm{n}_i + \bm{\beta}_i - \bm{\beta} \\
    R_{\bm{e}, j}(\bm{s}, \bm{n}_j) &= (\bm{\alpha}_j - \bm{\alpha})\bm{s} + \bm{\alpha}_j\bm{n}_j + \bm{\beta}_j - \bm{\beta}
\end{align*}
There are two cases.
If $\bm{\alpha}_i \not=\bm{\alpha}$, then $R_{\bm{e}, i}(\bm{s}, \bm{n}_i)$ is not a constant function of $\bm{s}$.
But if $\bm{\alpha}_i =\bm{\alpha}$, then $\bm{\alpha}_j \not=\bm{\alpha}$ and so $R_{\bm{e}, j}(\bm{s}, \bm{n}_j)$ is not a constant function of $\bm{s}$.
This is a contradiction, and so $\bm{\alpha}_i = \bm{\alpha}_j$ for all $i,j$.

Suppose similarly that there exist $\bm{\beta}_i \not=\bm{\beta}_j$.
If $\bm{\beta}_i \not=\bm{\beta}$, then $\mathbb{E}[R_{\bm{e}, i}(\bm{n}_i)] = \bm{\beta}_i - \bm{\beta}$ which is non-zero.
If $\bm{\beta}_i =\bm{\beta}$, then $\bm{\beta}_j \not=\bm{\beta}$ and so $\mathbb{E}[R_{\bm{e}, j}(\bm{n}_j)] = \bm{\beta}_j - \bm{\beta}$ is non-zero.
This is a contradiction, and so $\bm{\beta}_i = \bm{\beta}_j$ for all $i,j$.

We have thus proven that set $\{ \bm{e}_i \} \in \mathcal{G}_K$ is of the form $\bm{e}_i = \bm{\alpha} \bm{k}_i^{-1} + \bm{\beta}$ for all $i$.

\end{document}